\documentclass{article}
\usepackage[final]{corl_2026}

\usepackage[utf8]{inputenc} 
\usepackage[T1]{fontenc}    
\usepackage{hyperref}       
\usepackage{url}            
\usepackage{booktabs}       
\usepackage{amsfonts}       
\usepackage{nicefrac}       
\usepackage{microtype}  
\usepackage{xcolor}
\usepackage{graphicx}
\usepackage{svg}
\definecolor{cvprblue}{rgb}{0.21,0.49,0.74}
\definecolor{forestgreen}{RGB}{34,139,34}
\definecolor{winered}{RGB}{127,0,32}
\usepackage{amsmath} %
\usepackage{amssymb}  %
\usepackage{array}
\usepackage{multirow} 
\usepackage{makecell}
\usepackage{pifont}
\newcommand{\cmark}{\ding{51}}
\newcommand{\xmark}{\ding{55}}
\usepackage[x11names,dvipsnames]{xcolor}
\usepackage{colortbl}
\usepackage{framed}
\usepackage{wrapfig}  
\usepackage{titletoc}
\usepackage{color} 
\usepackage{caption, subcaption, overpic, textpos}
\usepackage{comment}
\usepackage{fontawesome}
\usepackage{pifont}
\usepackage{float}
\usepackage{booktabs}
\usepackage{siunitx}
\usepackage{wrapfig}
\newcommand{\Stars}[1]{\ding{72}\ifnum#1>1\repeatstar{#1}\fi}
\newcommand{\repeatstar}[1]{\ifnum#1>1\ding{72}\repeatstar{\numexpr#1-1}\fi}

\title{World In Your Hands: A Large-scale Ego-centric Dataset
for Learning Robotic Manipulation In the Wild}

\author{
\small
Yupeng Zheng$^{1,2,*}$, Jichao Peng$^{3,*}$, Weize Li$^{4,*}$,
Yuhang Zheng$^{4}$, Xiang Li$^{5}$,
Yujie Jin$^{3}$, Julong Wei$^{6}$, \\ Guanhua Zhang$^{3}$,
Ruiling Zheng$^{3}$, Ming Cao$^{3}$,
Songen Gu$^{3}$, Zhenhong Zou$^{3}$, Kaige Li$^{3}$,\\
Ke Wu$^{6}$, Mingmin Yang$^{3}$,
Jiahao Liu$^{3}$, Pengfei Li$^{3}$, Hengjie Si$^{3}$,
Feiyu Zhu$^{3}$, Wang Fu$^{3}$,\\
Likun Wang$^{3}$, Ruiwen Yao$^{3}$, Jieru Zhao$^{7}$,
Yilun Chen$^{3}$, Wenchao Ding$^{3}$\\[0.5em]
$^{1}$Institute of Automation, Chinese Academy of Sciences\\
$^{2}$University of Chinese Academy of Sciences
$^{3}$TARS Robotics \\
$^{4}$National University of Singapore
$^{5}$Tsinghua University \\
$^{6}$Fudan University
$^{7}$Shanghai Jiao Tong University\\[0.3em]
$^{*}$Equal contribution
}

\begin{document}
\maketitle
 \begin{figure}[!ht]
 \centering
 \includegraphics[width=0.9\textwidth]{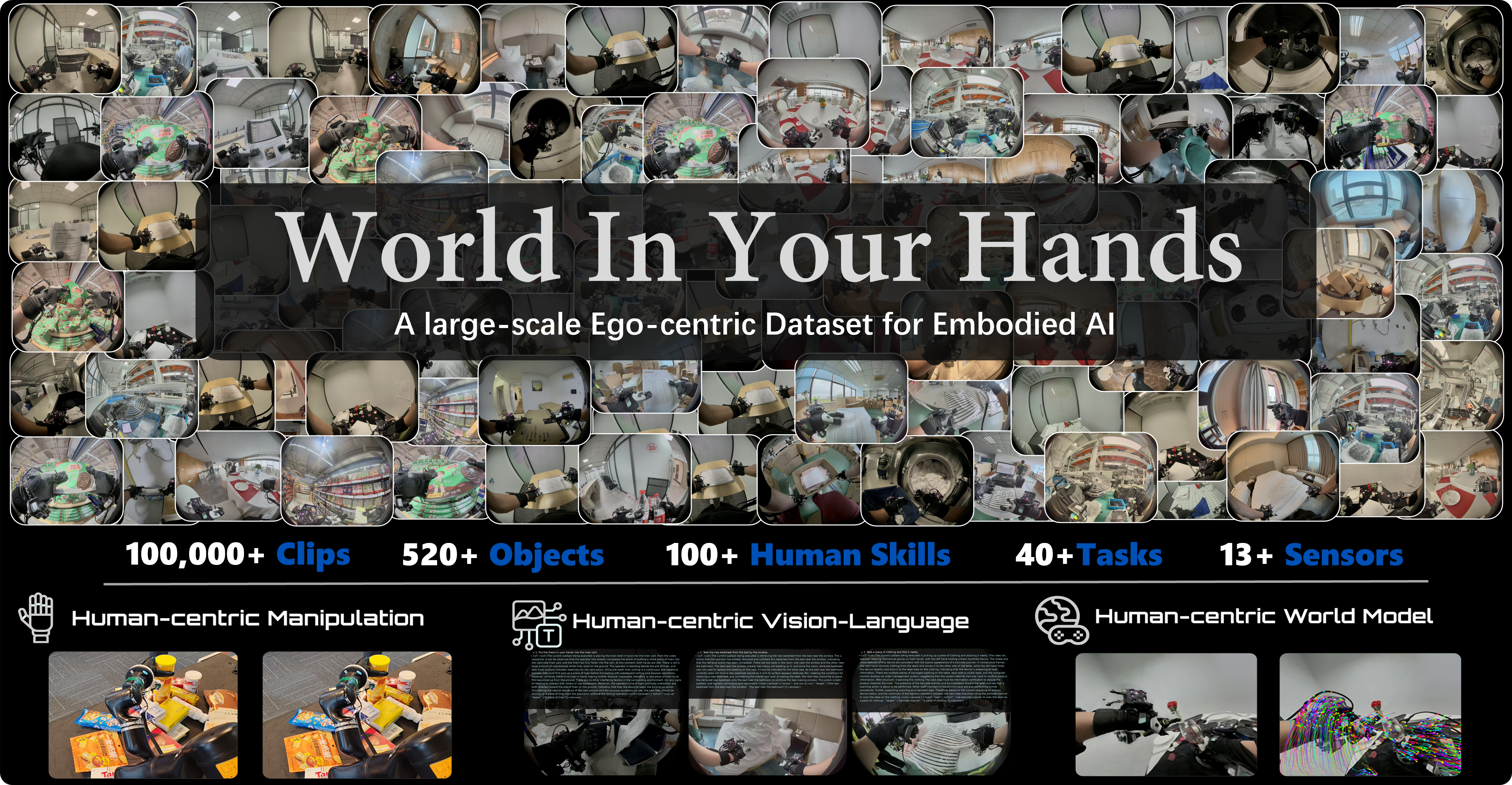}
\caption{\textbf{World In Your Hands (WIYH).} WIYH is a large-scale ego-centric dataset for embodied AI research, with a focus on learning robotic manipulation in the wild. It features diverse real-world human manipulation activities across household, service, logistics, and industrial environments. WIYH provides synchronized multimodal recordings and annotations, including RGB observations, 3D hand and wrist trajectories, hand skeletons, depth, masks, and language supervision, covering 40+ tasks, 100+ manipulation skills, and 500+ objects.}
 \label{fig:teaser}
 \end{figure}

\begin{abstract}
In this paper, we introduce World In Your Hands (WIYH), a large-scale ego-centric dataset comprising over 1,000 hours of in-the-wild human manipulation demonstrations with millimeter-level 3D motion accuracy. WIYH is constructed through an integrated data-construction pipeline that combines scalable acquisition with Oracle Suite, multi-stage action-language-perception annotation, and validation-based quality control. The dataset spans hundreds of manipulation skills and diverse real-world scenarios, with multimodal annotations including calibrated visual observations, 3D hand and wrist trajectories, hand skeletons, depth, masks, task and sub-task language, and vision-language supervision. We evaluate WIYH through ego-robot co-training and pre-training studies in RoboCasa simulation and real robots. Results show that ego-centric data improves manipulation policies when robot demonstrations are accurate but narrow, while verified WIYH annotations provide substantially stronger pre-training signals than unverified ego-centric data, even when the latter contains more samples. These findings highlight the importance of accurate action labeling and annotation quality control for turning scalable ego-centric data into effective robot-learning supervision. The dataset and documentation will be publicly available.
\end{abstract}

\keywords{Ego-centric Data, Robot Manipulation, Policy Pre-training}


\section{Introduction}
Large-scale pre-training has become a key driver of generalization in Large Language Models~\cite{brown2020language, chowdhery2023palm, achiam2023gpt, touvron2023llama,yang2025qwen3}, Vision-Language Models~\cite{radford2021learning,li2022blip,alayrac2022flamingo,li2023blip,li2024llava} and Vision-Language-Action (VLA) Models~\cite{brohan2022rt,zitkovich2023rt,kim2024openvla,black2410pi0,wen2024diffusion,zhou2025vision}. In robotic manipulation, however, scalable data collection remains a central bottleneck: robot demonstrations are costly to collect, often limited in scene diversity, and difficult to scale across objects, skills, and long-horizon workflows. Ego-centric human data offers a promising alternative, as it captures natural object interactions from the operator's viewpoint and can be collected in everyday environments at much larger scale. Existing efforts either use instrumented capture systems, such as VR devices~\cite{fan2025emhi} or smart glasses~\cite{punamiya2026egoverse}, to record human hand-object interactions~\cite{chao2021dexycb,kwon2021h2o,banerjee2025hot3d}, or extract action-related supervision from large-scale unlabeled ego-centric videos using foundation models~\cite{grauman2022ego4d,xu2021videoclip,wang2021actionclip,he2024egovm,li2025scalable}. Yet these approaches still struggle to jointly provide in-the-wild diversity, accurate 3D action supervision, aligned multimodal annotations, and evidence of usefulness for downstream robot policy learning.

\begin{table*}[b]
    \centering
    \setlength{\tabcolsep}{0.03\linewidth}
    \captionsetup{width=\textwidth}
    \caption{\textbf{Comparison to existing datasets.} 
    WIYH combines large-scale in-the-wild collection, calibrated multimodal observations, 3D action supervision, and language annotations (please see Appendix~\ref{app:dataset} for details), making it suitable for studying ego-centric data for robot policy learning.
    \label{tab:dataset_comparison} 
    \vspace{-2mm}
    } 

    \resizebox{\textwidth}{!}{%
        \begin{tabular}{l c c c c c c c c c c l}
            \toprule
            Dataset & Hours & Clips & Wild & \makecell{Cam. Calib} & RGB & Action & Depth & Mask & Instruct. & VLM \\
            \midrule
            Ego4D~\cite{grauman2022ego4d} & 3,670  & \textcolor{forestgreen}{\cmark} & \textcolor{forestgreen}{\cmark} & \textcolor{winered}{\xmark} & \textcolor{forestgreen}{\cmark} & 2D & \textcolor{winered}{\xmark} & \textcolor{winered}{\xmark} & \textcolor{forestgreen}{\cmark} & Subset \\
            HOI4D~\cite{liu2022hoi4d}     & 44.4  & 4k & \textcolor{forestgreen}{\cmark} & \textcolor{forestgreen}{\cmark} & \textcolor{forestgreen}{\cmark} &  2D & \textcolor{forestgreen}{\cmark} & \textcolor{forestgreen}{\cmark} & \textcolor{winered}{\xmark} & \textcolor{winered}{\xmark} \\
            HoloAssist~\cite{wang2023holoassist} & 166 & 2.22k & \textcolor{winered}{\xmark} & \textcolor{winered}{\xmark} & \textcolor{forestgreen}{\cmark} & - & \textcolor{forestgreen}{\cmark} & \textcolor{winered}{\xmark} & \textcolor{forestgreen}{\cmark} & Subset \\
            Ego-Exo4D~\cite{grauman2024ego} & 1,286 & 5.04k & \textcolor{forestgreen}{\cmark} & \textcolor{forestgreen}{\cmark} & \textcolor{forestgreen}{\cmark} & \textcolor{winered}{\xmark} & \textcolor{winered}{\xmark} & \textcolor{forestgreen}{\cmark} & \textcolor{forestgreen}{\cmark} & \textcolor{winered}{\xmark} \\
            CaptainCook4D~\cite{peddi2024captaincook4d} & 94.5 & 384 & \textcolor{forestgreen}{\cmark} & \textcolor{winered}{\xmark} & \textcolor{forestgreen}{\cmark} & \textcolor{winered}{\xmark} & \textcolor{forestgreen}{\cmark} & \textcolor{winered}{\xmark} & \textcolor{forestgreen}{\cmark} & \textcolor{winered}{\xmark} \\
            HOT3D~\cite{banerjee2025hot3d} & 13.9 & 425 & \textcolor{winered}{\xmark} & \textcolor{forestgreen}{\cmark} & \textcolor{forestgreen}{\cmark} & 3D & \textcolor{winered}{\xmark} & \textcolor{winered}{\xmark} & \textcolor{winered}{\xmark} & \textcolor{winered}{\xmark} \\
            HO-Cap~\cite{wang2024ho} & -  & 64 & \textcolor{winered}{\xmark} & \textcolor{forestgreen}{\cmark} & \textcolor{forestgreen}{\cmark} & 3D & \textcolor{forestgreen}{\cmark} & \textcolor{forestgreen}{\cmark} & \textcolor{winered}{\xmark} & \textcolor{winered}{\xmark} \\
            Ego-ViD-5M~\cite{wang2024egovid} & - & 5M & \textcolor{winered}{\xmark} & \textcolor{forestgreen}{\cmark} & \textcolor{forestgreen}{\cmark} & \textcolor{winered}{\xmark}  & \textcolor{winered}{\xmark} & \textcolor{winered}{\xmark} & \textcolor{forestgreen}{\cmark} & \textcolor{winered}{\xmark} \\
            EgoDex~\cite{hoque2025egodex} & 829 & 338k & \textcolor{forestgreen}{\cmark} & \textcolor{winered}{\xmark} & \textcolor{forestgreen}{\cmark} &  3D & \textcolor{winered}{\xmark} & \textcolor{forestgreen}{\cmark} & \textcolor{forestgreen}{\cmark} & \textcolor{winered}{\xmark} \\
            VITRA~\cite{li2025scalable} & - & 1M & \textcolor{forestgreen}{\cmark} & \textcolor{forestgreen}{\cmark} & \textcolor{forestgreen}{\cmark} & 3D & \textcolor{forestgreen}{\cmark} & \textcolor{winered}{\xmark} & \textcolor{forestgreen}{\cmark} & \textcolor{winered}{\xmark} \\
            \midrule
            \textbf{WIYH (Ours)} & 1045 & 125.4k & \textcolor{forestgreen}{\cmark} & \textcolor{forestgreen}{\cmark} &\textcolor{forestgreen}{\cmark} & 3D & \textcolor{forestgreen}{\cmark} & \textcolor{forestgreen}{\cmark} & \textcolor{forestgreen}{\cmark} & Subset \\
            \bottomrule \\
        \end{tabular}
    }
    \vspace{-5mm}
    \addtolength{\tabcolsep}{2pt}
\end{table*}

Despite these advances, existing ego-centric manipulation datasets still face three major limitations for robotic manipulation learning, as summarized in Table~\ref{tab:dataset_comparison}: 
\textbf{(a) Limited scenario diversity.} Many datasets are collected in constrained laboratory environments, limiting their coverage of real-world scene diversity, object variability, and long-horizon workflows. 
\textbf{(b) Inadequate vision-language-action alignment.} Existing collections often lack one or more key forms of aligned supervision, such as calibrated observations, accurate 3D hand or wrist motion, object-level masks, or fine-grained task language. This makes it difficult to directly use them for learning manipulation policies that require both executable actions and semantic task understanding. 
\textbf{(c) Insufficient policy-level evaluation.} Most datasets are evaluated mainly through perception, reconstruction, or recognition tasks, leaving it unclear whether their ego-centric annotations provide effective supervision for downstream robot policy learning.


To bridge this gap, we introduce World In Your Hands (WIYH), a large-scale ego-centric dataset for learning robotic manipulation in the wild, as shown in Figure~\ref{fig:teaser}. WIYH is designed to provide not only large-scale human manipulation demonstrations, but also accurate action supervision, aligned multimodal annotations, and quality control for downstream robot policy learning. Specifically, WIYH makes three dataset-level contributions:

\begin{itemize}
    \item \textbf{Oracle Suite.} We introduce a flexible wearable data-collection kit for scalable in-the-wild ego-centric capture. Oracle Suite records multimodal observations and produces 3D wrist trajectories with millimeter-level translational accuracy, without requiring external third-person tracking infrastructure.
    \item \textbf{Large-scale Demonstration Collection.} We present over 1,000 hours of multimodal human manipulation demonstrations across 10 diverse real-world scenarios. The collection includes multi-view RGB, depth, camera calibration, hand skeletons, and 3D wrist trajectories, covering a broad range of objects, skills, and long-horizon manipulation activities.
    \item \textbf{Quality-controlled Annotations.} We provide a multi-stage annotation and validation pipeline that produces 600 hours, or 100K episodes, of atomic action instructions. These annotations align accurate 3D actions with task language and perception supervision, enabling policy learning and analysis of how annotation quality affects downstream performance.
\end{itemize}

To evaluate the utility of WIYH for robot learning, we conduct two sets of policy-level studies. First, we perform a controlled real-robot co-training study in which accurate but narrowly collected robot demonstrations are combined with diverse ego-centric demonstrations of the same task. This study tests whether ego-centric data can expand the observation and action coverage available to a robot policy. Second, we evaluate ego-centric pre-training in RoboCasa Benchmark and real robots, comparing WIYH with existing ego-centric datasets and an unverified WIYH variant, WIYH*. Across these studies, WIYH improves downstream manipulation performance, while WIYH* provides weaker or inconsistent gains due to lower-quality action labels. These results show that the value of ego-centric data depends not only on scale and visual diversity, but also on accurate action labeling, annotation verification, and aligned language supervision.

\section{Related Works}
\label{sec:related}

\noindent \textbf{Ego-centric manipulation datasets.}
Large-scale ego-centric datasets have advanced video understanding and human-object interaction research. Ego4D~\cite{grauman2022ego4d} and Ego-Exo4D~\cite{grauman2024ego} provide broad first-person or multi-view recordings of daily activities, while HOI4D~\cite{liu2022hoi4d}, HoloAssist~\cite{wang2023holoassist}, and CaptainCook4D~\cite{peddi2024captaincook4d} focus on human-object interaction, assistance, and procedural activities. Closer to manipulation, DexYCB~\cite{chao2021dexycb}, H2O~\cite{kwon2021h2o}, HOT3D~\cite{banerjee2025hot3d}, HO-Cap~\cite{wang2024ho}, and EgoDex~\cite{hoque2025egodex} provide calibrated sensing, 3D hand or object motion, depth, masks, or richer interaction annotations. However, existing datasets rarely combine large-scale in-the-wild collection, accurate 3D action supervision, aligned language-perception annotations, and policy-level evaluation. WIYH is designed to fill this gap with quality-controlled 3D actions and multimodal annotations for downstream robot learning.

\noindent \textbf{Robot learning from ego-centric videos and demonstrations.}
Ego-centric data has recently been explored as a scalable source of supervision for robot learning. One line of work leverages large-scale unlabeled ego-centric videos to learn visual-language representations, action priors, affordance cues, or embodied reasoning signals for downstream policies~\cite{xu2021videoclip,wang2021actionclip,he2024egovm}. Another line uses curated ego-centric demonstrations with explicit hand, wrist, or object motion as cross-embodiment supervision for policy learning and world modeling~\cite{hsieh2025dexman,zhu2025learning,routray2025vipra,zhu2025let,kareer2025egomimic}. While these studies suggest that ego-centric data can improve robustness, generalization, and scene understanding, it remains unclear which properties of ego-centric supervision are most critical for robot manipulation. WIYH addresses this question from a dataset perspective by combining large-scale in-the-wild demonstrations, quality-controlled 3D actions, and aligned language-perception annotations, with evaluations that compare verified and unverified ego-centric supervision for downstream policy learning.

\section{\textit{World In Your Hands} Dataset Construction}
\label{sec:ecosystem}
We construct WIYH through a unified pipeline designed to collect, annotate, and validate large-scale ego-centric manipulation demonstrations in real-world environments. The pipeline includes multimodal acquisition with Oracle Suite, multi-stage annotation for action-language-perception alignment, validation and quality control for motion labels, and dataset-level statistical analysis.

\subsection{Data Acquisition with Oracle Suite}
\begin{figure*}[h!]
  \centering
  \setlength{\abovecaptionskip}{0.cm}
  \includegraphics[width= 0.85\textwidth]{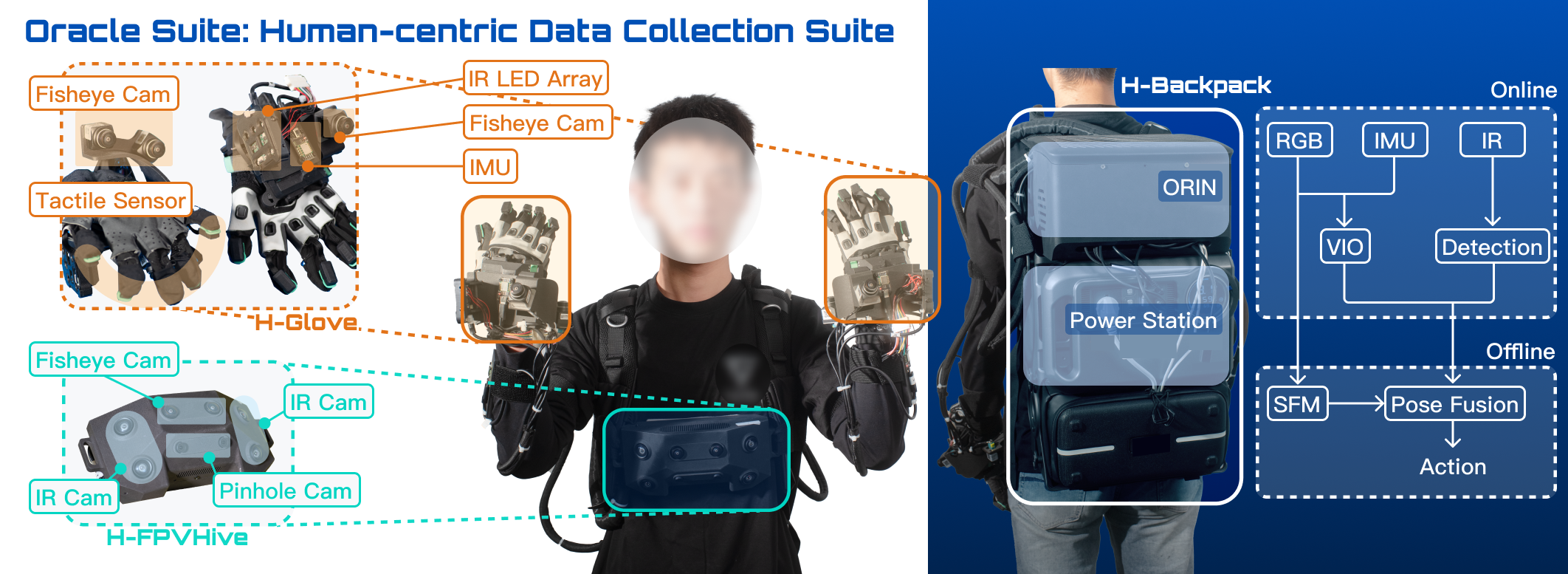}
  \vspace{1mm}
  \caption{\textbf{Oracle Suite for ego-centric data collection.} Oracle Suite consists of three synchronized wearable modules: \textbf{H-FPVHive}, a first-person perception module with multi-view cameras and infrared sensing; \textbf{H-Gloves}, hand motion-capture modules combining Manus~\cite{manusglove} gloves, IMUs, and visual trackers; and \textbf{H-Backpack}, an onboard computation, storage, and power unit. Together, these modules support in-the-wild multimodal recording and automatic 6D wrist-pose labeling.}
  \label{fig:suite}
\end{figure*}

To enable scalable acquisition of high-fidelity ego-centric manipulation data in unconstrained environments, we develop \textbf{Oracle Suite}, a portable wearable system for multimodal data collection. As illustrated in Figure~\ref{fig:suite}, Oracle Suite integrates synchronized wearable sensing with an automatic motion-labeling pipeline. During data collection, operators wear the system and perform manipulation tasks following scenario-specific standard operating procedures, while Oracle Suite records first-person visual observations, hand-centric observations, IMU measurements, and synchronized timestamps for later annotation and validation.

\textbf{Wearable modules.}
Oracle Suite consists of three primary modules: H-FPVHive, H-Gloves, and H-Backpack. H-FPVHive is a chest-mounted perception module equipped with two fisheye cameras, two pinhole cameras, and four infrared modules. These sensors are hardware-synchronized to capture the operator's first-person visual context and support relative localization of the H-Gloves. Each H-Glove integrates a Manus glove, six IMUs, an onboard MCU, and three fisheye cameras for hand motion sensing and visual tracking. H-Backpack provides onboard computation, data storage, and power through an NVIDIA Orin computing unit.

\textbf{Automatic motion labeling.}
The automatic labeling pipeline combines online synchronization with offline motion reconstruction. By fusing infrared, RGB, and IMU measurements, the pipeline estimates 6D wrist pose trajectories for both hands and aligns them with the recorded visual streams. These trajectories provide executable 3D action supervision for downstream robot learning, while the detailed validation and filtering procedures are described in Section~3.3.

\textbf{Collection efficiency and tracking robustness.}
Oracle Suite is designed for long-duration in-the-wild data collection without fixed laboratory tracking infrastructure or extensive operator training. In our workflow, Oracle Suite achieves approximately 5$\times$ higher collection throughput than teleoperated robot data collection, increasing the daily collection rate from 150 to 720 episodes. Compared with VR-based hand tracking, Oracle Suite provides more reliable 3D hand skeletons under visual occlusion, as validated in our tracking comparison study. These properties make Oracle Suite suitable for scaling ego manipulation data collection across diverse scenes, operators, objects and skills.

\begin{figure}[h!]
  \centering
  \setlength{\abovecaptionskip}{0.cm}
  \includegraphics[width= 0.85\textwidth]{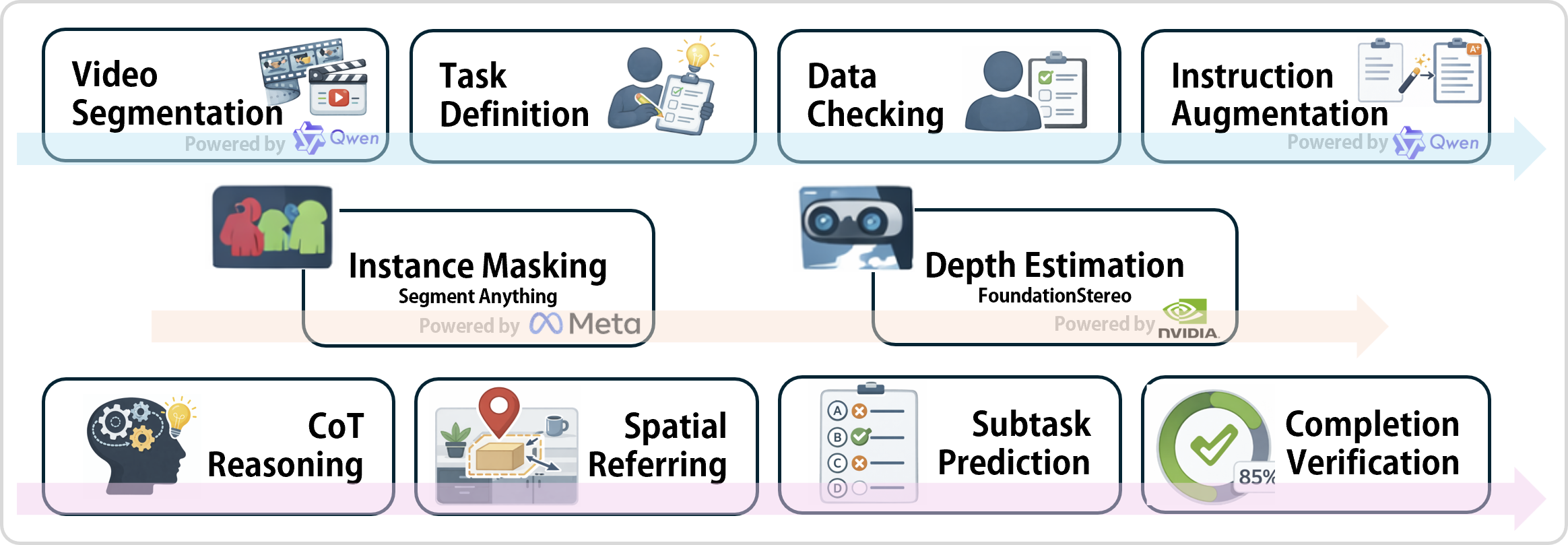}
  \vspace{1mm}
  \caption{\textbf{Overview of the WIYH annotation pipeline.} Synchronized raw recordings are converted into aligned action-language-perception supervision through \textcolor[HTML]{B0D2DE}{atomic action annotation}, \textcolor[HTML]{F9D4BD}{perception annotation}, and \textcolor[HTML]{EEC0E9}{vision-language annotation}. Automatically generated segments and labels are manually reviewed to ensure annotation consistency and quality.}
  \label{fig:data_pipeline}
\end{figure}

\subsection{Data Annotation Pipeline}\label{sec:data_collection_annotation}

After data acquisition, we convert synchronized raw recordings into aligned supervision for robot learning through a three-stage annotation pipeline, as shown in Figure~3. The pipeline produces atomic action instructions, perception annotations, and vision-language annotations, followed by human verification to ensure consistency across action, language, and visual observations.

\textbf{Atomic action annotation.}
Long-horizon demonstrations are first decomposed into semantically meaningful atomic segments. We use Qwen2.5-VL-72B~\cite{Qwen2.5-VL} to propose temporal boundaries for each recorded task, where each segment corresponds to a basic manipulation unit. Human annotators then verify the segment boundaries and assign each segment a preset manipulation skill and target object. Finally, the skill and object labels are converted into atomic task instructions and further augmented with Qwen2.5-VL-72B to improve linguistic diversity while preserving the action semantics.

\textbf{Perception annotation.}
For each validated segment, we associate the recorded motion labels with calibrated visual observations and perception annotations, including camera parameters, depth maps, object or hand masks, 3D wrist trajectories, and hand skeletons. These annotations provide the visual and geometric context required for aligning human actions with robot-learning observations.

\textbf{Vision-language annotation.}
Beyond atomic instructions, we further annotate a subset of data with higher-level vision-language supervision, including task descriptions, sub-task descriptions, and reasoning-oriented annotations for manipulation-centric understanding. These annotations are intended to support broader embodied AI tasks, while our main experiments focus on policy learning.

\textbf{Human verification.}
All automatically generated temporal segments and language annotations are manually reviewed. Annotators correct boundary errors, skill-object mismatches, and inconsistent instructions, producing quality-controlled action-language-perception alignment for downstream training and evaluation.

\subsection{Data Validation and Quality Control}
\label{sec:check}
\begin{figure*}[]
  \centering
  \setlength{\abovecaptionskip}{0.cm}
  \includegraphics[width= 0.8\textwidth]{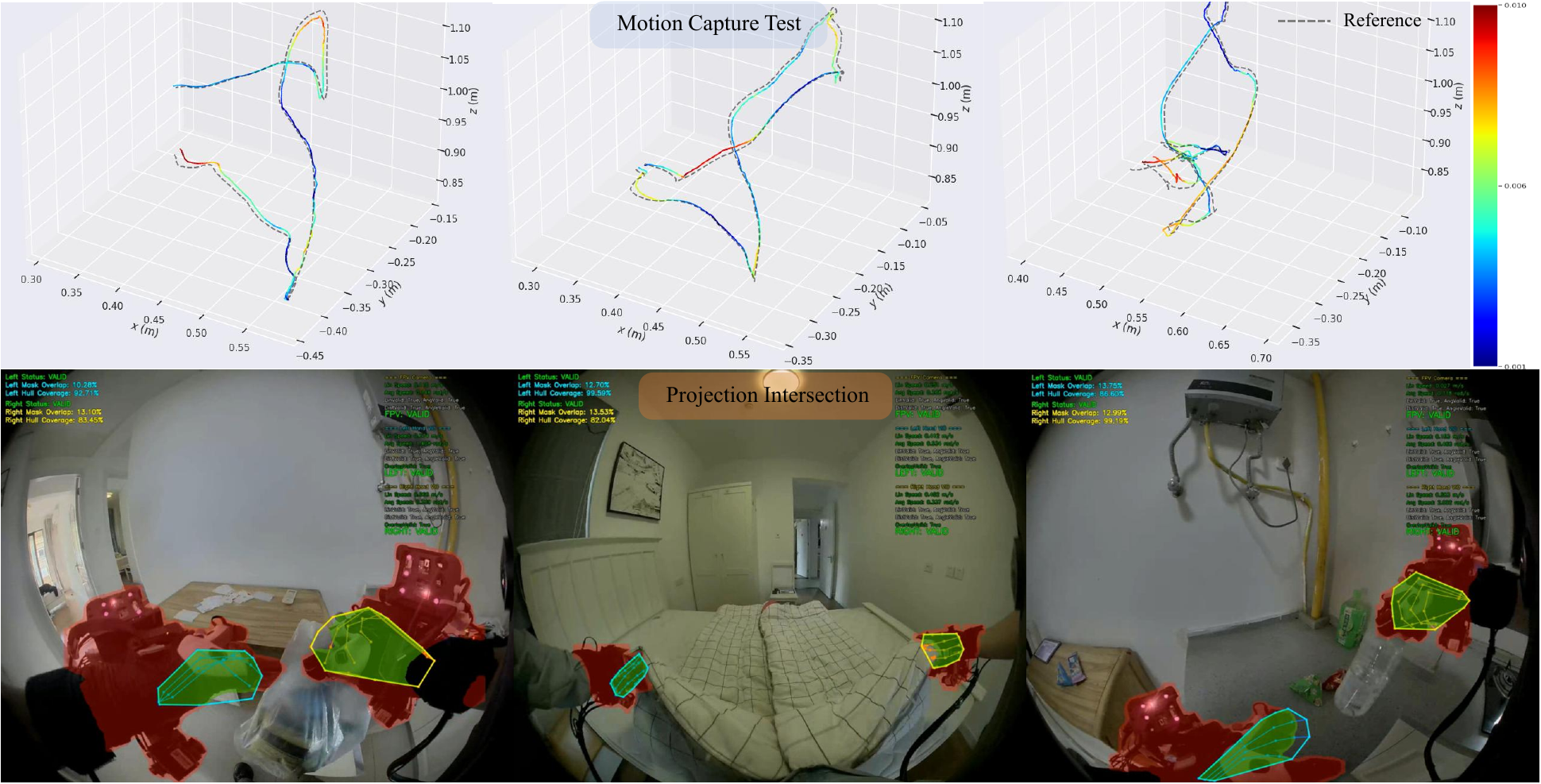}
  \vspace{1mm}
  \caption{\textbf{Oracle Suite Accuracy and Dataset Quality Control.} Top: controlled motion-capture validation comparing Oracle Suite wrist-pose estimates with reference trajectories, where color indicates positional deviation. Bottom: reprojection-based quality checks for in-the-wild data, where estimated hand trajectories and skeletons are projected into first-person views and compared with segmented hand masks.}
  \vspace{-4mm}
  \label{fig:dataset_check}
\end{figure*}

To ensure reliable action supervision, we validate WIYH at two levels. First, as shown in Figure~\ref{fig:dataset_check}, we evaluate Oracle Suite in a controlled motion-capture room by attaching reflective markers to the H-Gloves and comparing estimated wrist trajectories with motion-capture references. Oracle Suite achieves a mean positional error below 5~mm, supporting its use for generating high-accuracy 3D wrist action labels. Second, for large-scale in-the-wild data where motion-capture ground truth is unavailable, we use reprojection-based consistency checks: estimated hand skeletons and wrist trajectories are projected into calibrated first-person views and compared with segmented hand masks. Segments with poor projection consistency are flagged for manual review or removed. Together with human checks of temporal boundaries and task definitions, these procedures form the quality-control mechanism used to construct the verified WIYH dataset.



\subsection{Dataset Statistics}
\begin{figure*}[b]
  \centering
  \setlength{\abovecaptionskip}{0.cm}
  \includegraphics[width=0.80 \textwidth]{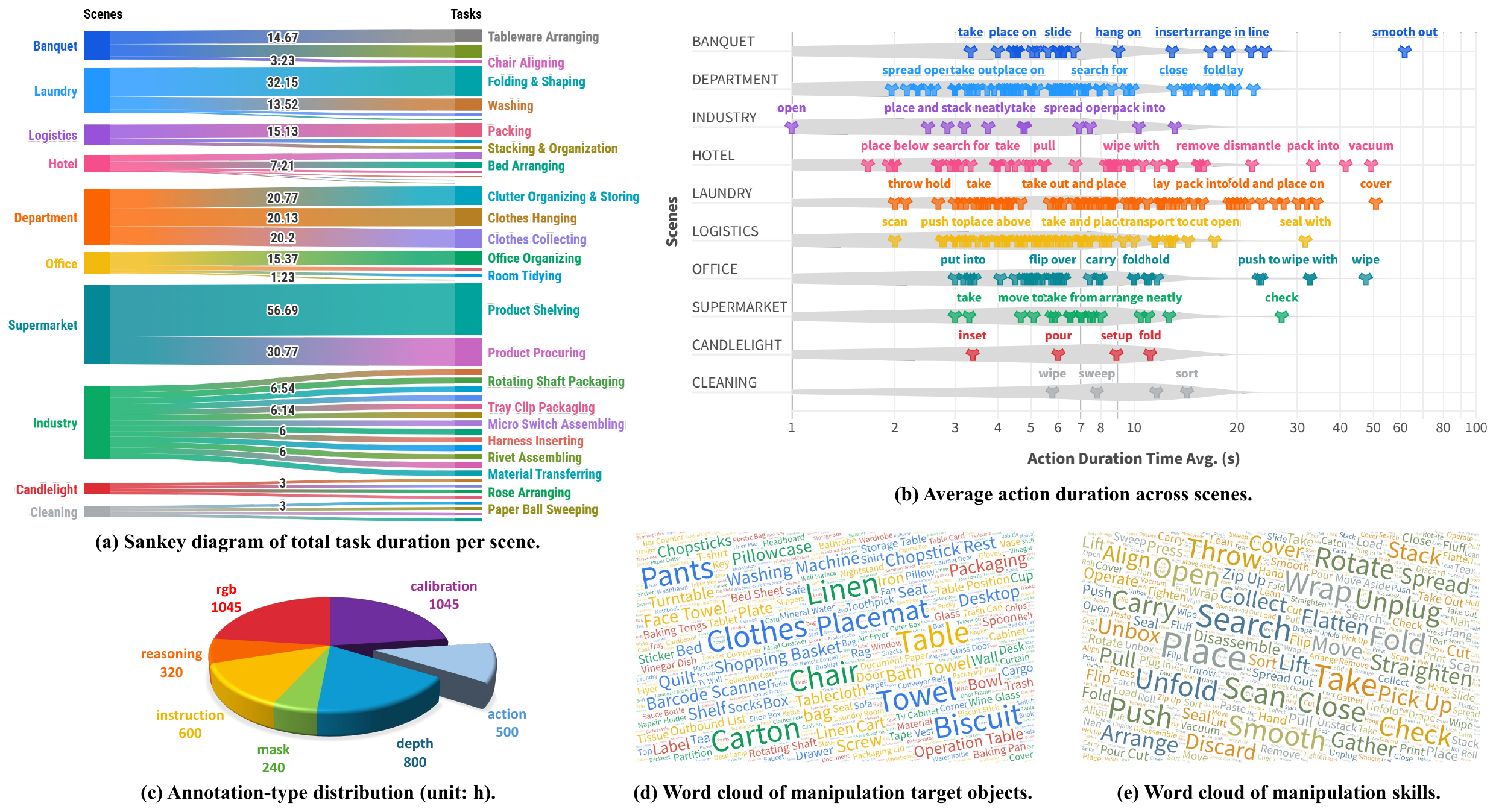}
  \caption{Dataset statistics of WIYH. (a) Task-duration distribution across real-world scenarios. (b) Average duration of atomic actions across scenarios. (c) Annotation coverage in hours. (de) Word clouds of annotated target objects and manipulation skills. These statistics demonstrate the scenario, temporal, object, skill, and annotation diversity of WIYH.} 
  \label{fig:dataset statistics}
  \vspace{-2mm}
\end{figure*}

Figure~\ref{fig:dataset statistics} summarizes the scale, diversity, and annotation coverage of WIYH. The dataset contains 1,045 hours of in-the-wild ego-centric manipulation demonstrations across 10 real-world scenarios, including banquet, laundry, logistics, hotel, department store, office, supermarket, industrial, cleaning, and candlelight settings. As shown in Figure~\ref{fig:dataset statistics}(a), these scenarios cover a wide range of task durations and workflows, from short structured routines to long-horizon household and industrial activities. This scenario-level diversity is important for exposing policies to varied environments, object layouts, and human manipulation strategies.
Figure~\ref{fig:dataset statistics}(b) analyzes the temporal distribution of atomic actions. WIYH contains both short actions, such as wiping, inserting, and placing, and longer-horizon actions, such as smoothing, folding, and arranging. These temporal variations reflect the heterogeneous structure of real-world manipulation, where policies must reason over both fine-grained contact events and extended task progress.
Figure~\ref{fig:dataset statistics}(c) reports the annotation coverage of WIYH. In addition to full RGB recordings and calibration, WIYH provides large-scale 3D action trajectories, depth, masks, and atomic instructions, with higher-level reasoning and vision-language annotations available for a subset of the data. This annotation structure supports both robot policy learning and broader manipulation-centric Embodied AI tasks.
Finally, Figure~\ref{fig:dataset statistics}(d)(e) visualizes the distributions of annotated target objects and manipulation skills. WIYH covers diverse object categories, including clothes, linens, cartons, cups, and tableware, spanning deformable, rigid, and articulated objects. Its skill vocabulary includes manipulation primitives such as take, place, rotate, unfold, align, push, and search. Together, these statistics show that WIYH provides broad coverage across scenarios, objects, skills, temporal scales, and annotation types. More data, demonstrations, and statistics will be provided in Appendix~\ref{app:dataset}.

\section{Does High-Quality Ego-Centric Data Help Robot Policy Learning?}
\subsection{Evaluation Protocol}
To answer whether high-quality ego-centric data helps robot policy learning, we evaluate WIYH through three complementary policy-level evaluations. The first study tests whether ego-centric demonstrations can complement accurate but narrowly collected robot demonstrations. In this setting, robot data provide reliable executions in limited scenes, while ego-centric data collected with Oracle Suite provides broader observation and action coverage. The second study evaluates ego-centric pre-training followed by robot-data post-training in RoboCasa simulation and on real robots. This setting tests whether ego-centric data provides transferable manipulation supervision across embodiment, observation, and action-domain gaps. Across both studies, we focus on dataset-level questions rather than proposing a new policy architecture: when does ego-centric data help, and how much does annotation quality matter?



\textbf{Task-matched ego-robot co-training.}
We construct a combined training set from robot demonstrations collected in constrained scenes and ego-centric demonstrations of the same task collected with Oracle Suite. Robot data provide accurate executions from limited initial poses, while ego-centric data is collected in cluttered multi-object scenes and provides broader observation and action coverage. We evaluate generalization in single-object and cluttered multi-object settings to test whether ego-centric data complements accurate but narrowly collected robot demonstrations. Details of the robot hand, retargeting procedure, and evaluation settings are provided in Appendix~\ref{app:cotrain_details}.

\textbf{RoboCasa pre-training with matched-scale ego data.}
We compare ego-centric pre-training using EgoDex, VITRA, WIYH, and WIYH*, followed by post-training on 10\% of the official RoboCasa demonstrations. WIYH* is an unverified variant of WIYH that removes the reprojection-based filtering and manual task-segmentation verification described in Section~\ref{sec:check}. To isolate annotation quality from scale, all pre-training datasets are matched to 80K episodes. We use Fast-WAM~\cite{yuan2026fast} as the base model; training details and the RoboCasa evaluation protocol are provided in Appendix~\ref{app:robocasa_details}.

\textbf{WIYH pre-training for real-robot transfer.}
We further evaluate whether WIYH provides transferable supervision for physical robot manipulation by pre-training on WIYH and post-training on two real-world tasks: rose insertion and gift packing. We compare policies trained with and without WIYH pre-training under varied object positions and seen/unseen target objects. Data collection and evaluation details are provided in Appendix~\ref{app:realrobot_details}.
\begin{figure}[]
  \centering
  \setlength{\abovecaptionskip}{0.cm}
  \includegraphics[width=0.95\textwidth]{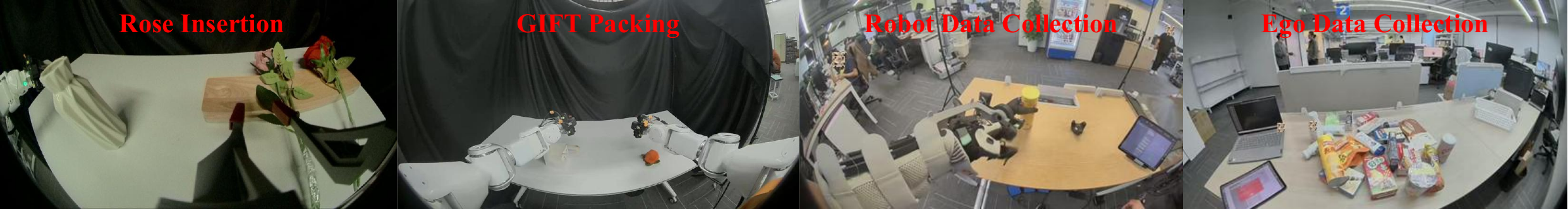}
  \caption{\textbf{Experiment setup.} Left: real-robot transfer tasks after WIYH pre-training, including rose insertion and gift packing. Right: task-matched ego-robot co-training setup, where robot data are collected in single-object scenes with limited initial poses, while ego-centric data are collected in cluttered multi-object scenes with broader observation and action coverage.}
  \label{fig:setup}
  \vspace{-4mm}
\end{figure}

\subsection{Results and Analysis}
We analyze the results of the task-matched ego-robot co-training study, the RoboCasa pre-training study, and the real-robot transfer study. Together, these results evaluate whether WIYH provides useful supervision for robot policy learning and how annotation quality affects downstream performance.

\textbf{Task-matched ego-robot co-training.}
Table~\ref{tab:real_exp} summarizes the co-training results. In the single-object setting, adding ego-centric demonstrations improves the success rate from 43.3\% to 56.7\% when using 500 robot clips, yielding a 13.4\% gain. This suggests that ego-centric data helps the policy generalize to unseen initial robot poses. The benefit is larger in the cluttered multi-object setting:
\begin{wraptable}{r}{0.50\linewidth}
\vspace{-1.0em}
\centering
\caption{Task-matched ego-robot co-training results. Each row reports robot-only training versus adding 800 ego-centric clips.}
\vspace{-0.6em}
\resizebox{1.00\linewidth}{!}{
\begin{tabular}{lcccc}
\toprule
\textbf{Scenario} & \textbf{Robot} & \textbf{Ego} & \textbf{SR (\%)} & \textbf{Gain} \\
\midrule
Single-object & 200 & 0 $\rightarrow$ 800 & 20.0 $\rightarrow$ 23.3 & \textbf{+3.3} \\
Single-object & 500 & 0 $\rightarrow$ 800 & 43.3 $\rightarrow$ 56.7 & \textbf{+13.4} \\
Cluttered & 200 & 0 $\rightarrow$ 800 & 0.0 $\rightarrow$ 32.0 & \textbf{+32.0} \\
Cluttered & 500 & 0 $\rightarrow$ 800 & 8.0 $\rightarrow$ 60.0 & \textbf{+52.0} \\
\bottomrule
\end{tabular}
}
\vspace{-1.2em}
\label{tab:real_exp}
\end{wraptable}
 Increasing robot data alone from 200 to 500 clips improves success only from 0.0\% to 8.0\%, whereas adding 800 ego-centric clips raises success to 60.0\%. These results show that when robot data are accurate but visually and behaviorally narrow, task-matched ego-centric data can substantially expand the observation and action coverage available for policy learning.

\begin{figure*}[]
  \centering
  \includegraphics[width=0.95\textwidth]{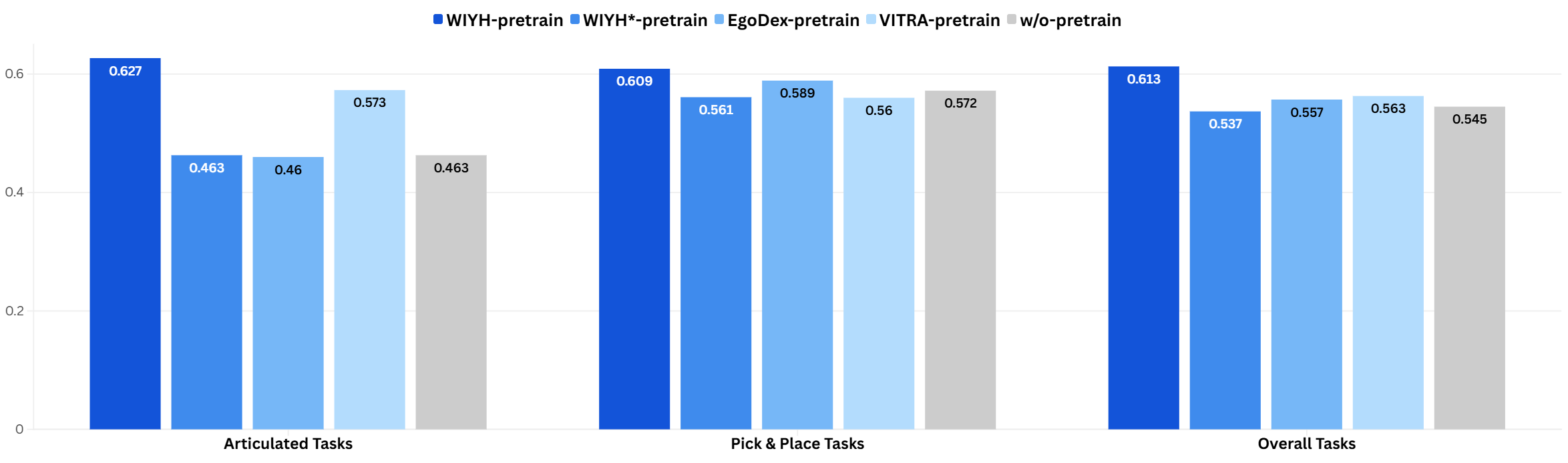}
  \caption{\textbf{RoboCasa Experiments.} We report results on the RoboCasa GR1 tabletop simulation using different ego-centric pre-training datasets. WIYH* denotes WIYH data without verification, which includes human-collected data with low-quality actions.}
  \vspace{-5mm}
  \label{fig:robocasa}
\end{figure*}

\textbf{RoboCasa pre-training with matched-scale ego data.}
Figure~\ref{fig:robocasa} compares different ego-centric pre-training datasets under matched data scale. WIYH achieves the strongest performance on articulated-object tasks and overall performance comparable to the best baseline, indicating that its verified 3D action labels and aligned annotations provide effective pre-training supervision. In contrast, WIYH* substantially underperforms verified WIYH and provides little overall gain over the no-pretraining baseline. Since WIYH* has the same data source but removes reprojection-based filtering and manual task-segmentation verification, this comparison suggests that noisy ego-centric action labels and unreliable action-language alignment can offset the benefits of ego-centric scale. The gains of WIYH are most pronounced on articulated-object tasks, where accurate 3D action labels and verified temporal segmentation are especially important. Pick-and-place tasks show smaller differences across pre-training settings, suggesting that they are less sensitive to the specific ego-centric pre-training distribution under the current post-training protocol. These results support the central claim that ego-centric data is most useful when scale is paired with accurate action labeling and annotation verification.



\textbf{WIYH pre-training for real-robot manipulation.}
Table~\ref{tab:real} reports the real-world results. Compared with training without pre-training, WIYH pre-training improves the success rate from 15\% to 70\% across rose insertion and gift packing. The improvement is especially clear on unseen objects, suggesting that WIYH provides transferable object-interaction priors beyond the limited post-training data. 
\begin{wraptable}{r}{0.54\linewidth}
\vspace{-1.0em}
\centering
\caption{We compare policies trained with and without WIYH pre-training on two real-robot tasks, including seen and unseen objects.}
\vspace{-0.6em}
\resizebox{1.00\linewidth}{!}{
\begin{tabular}{lccccc} 
\toprule
\multirow{2}{*}{Method} 
& \multicolumn{2}{c}{Rose Insertion} 
& \multicolumn{2}{c}{Gift Packing} 
& \multirow{2}{*}{AVG} \\
& Seen Bottle & Unseen Bottle & Seen Gift & Unseen Gift & \\
\midrule
w/o Pretrain & 1/4 & 0/6  & 1/4 & 1/6 & 15\% \\
w. Pretrain & 3/4 & 3/6 & 4/4 & 4/6 & 70\% \\
\bottomrule
\end{tabular}
}
\vspace{-1.0em}
\label{tab:real}
\end{wraptable}
This benefit is consistent with the composition of WIYH, which contains diverse packing and insertion interactions from logistics and candlelight-dinner scenarios. These results provide physical-robot evidence that WIYH can serve as useful pre-training supervision for manipulation policies.

\section{Conclusion and Limitations}
\label{sec:conclusion}
We introduced World In Your Hands (WIYH), a large-scale ego-centric dataset for learning robotic manipulation in the wild. WIYH provides over 1,000 hours of multimodal human manipulation demonstrations collected with Oracle Suite, together with quality-controlled annotations including calibrated visual observations, 3D hand and wrist trajectories, hand skeletons, depth, masks, task language, and vision-language supervision. Through ego-robot co-training, RoboCasa pre-training, and real-robot transfer experiments, we show that ego-centric data can complement accurate but narrow robot demonstrations and provide useful pre-training supervision for manipulation policies. Our verified-versus-unverified comparison further highlights that the value of ego-centric data depends not only on scale, but also on accurate action labeling and annotation quality control. As limitations, WIYH currently focuses on visual, geometric, language, and motion annotations; future extensions could incorporate full-hand tactile sensing and broader policy evaluations across more robot embodiments and manipulation categories.

\bibliography{main}  

\newpage
\appendix

\section{Oracle Suite: Hardware and Collection Efficiency}
\subsection{Hardware Details}
\textbf{Camera:} We primarily use two fisheye cameras below each wrist and two fisheye cameras on the chest as the raw RGB signal sources. These cameras feature a 180° field of view with an original resolution of 1536 × 1920 pixels.

\noindent\textbf{Autolabel Pipeline:} The autolabeling pipeline involves online and offline algorithms. Specifically, the online localization module performs lightweight coarse positioning using visual-inertial odometry (VIO) algorithms~\cite{qin2018robust} and IR detection, while the offline stage refines these coarse estimates through Structure-from-Motion (SfM) algorithms~\cite{schoenberger2016sfm}.

\subsection{Data Collection Efficiency}
The Oracle Suite pipeline follows a \textit{Collection, Annotation, Inspection, Archiving} workflow for scalable data production.  A single collector operating 8 hours per day generates approximately 1.8\,TB of valid multimodal data. At scale, 100 collectors over 30 days yield roughly 5.4\,PB of raw data (stored in cloud repositories). Efficiency is further improved through standardized capture protocols and automated cloud-based annotation.

We compare the efficiency of different data collection paradigms in Table~\ref{tab:collection}. Efficiency: taking the rose insertion task in our experiments as an example, given 8 hours as collection time, teleoperation can collect about 150 data clips; UMI-like methods, such as DexUMI~\cite{xu2025dexumi}, can collect about 400 data clips; our oracle suite can collect 720 data clips; and VR can collect around 800 data clips. 
Quality: VR-based hand tracking only provides 2D hand skeletons and suffers from visual occlusion. In contrast. In comparison, the Oracle Suite provides accurate 3D hand skeletons, even if under visual occlusion.

\begin{table}[h]
\setlength{\tabcolsep}{0.03\linewidth}
\centering
\caption{
\textbf{Comparison of dexterous data collection paradigms.}
}
\label{tab:collection}
\begin{center}
\resizebox{1.0\textwidth}{!}{
    \begin{tabular}{lccccc}
    \toprule
     & \textbf{Teleop} & \textbf{DexUMI} & \textbf{VR} & \textbf{Oracle Suit (Ours)} \\
    \midrule
    Efficiency & \Stars{1} & \Stars{2} & \Stars{5} & \Stars{4} \\
    Quality & \Stars{5} & \Stars{4} & \Stars{2} & \Stars{4} \\
    End-effector & Dexterous & Dexterous & Hands & Hands \\
    \bottomrule
    \end{tabular}
}
\end{center}
\end{table}

\section{WIYH Dataset Details}
\label{app:dataset}

\subsection{Task and Skill Statistics}
\begin{figure*}[ht]
  \centering
    \includegraphics[width=\textwidth]{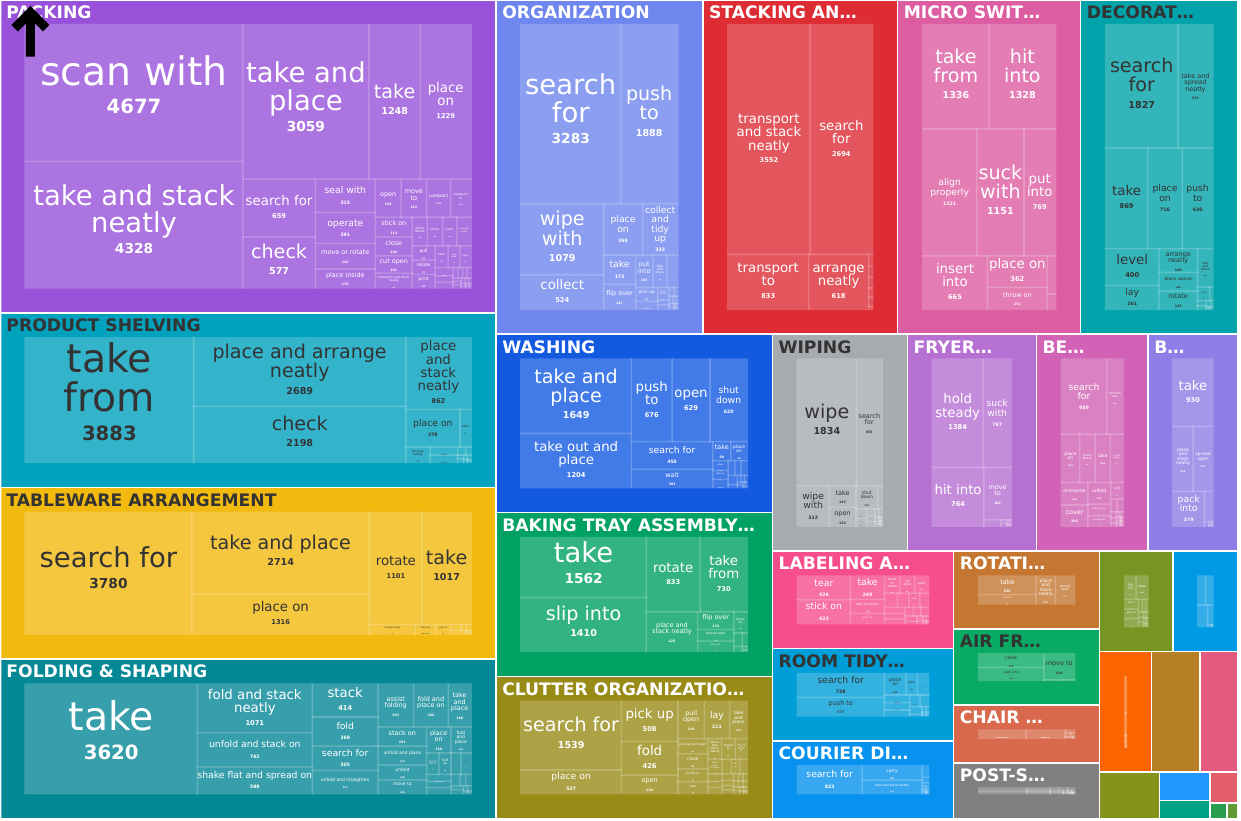}
  \caption{Task-skill count visualization of 25\% dataset.}
  \label{fig:vis_sta}
\end{figure*}

We further supplement the visualization of the number of skills corresponding to each task. Due to space limitations, we report statistics based on only 25\% of the dataset, as shown in Figure~\ref{fig:vis_sta}. Please zoom in electronically to inspect the details. The full visualization analysis will be updated subsequently on the dataset website.

\subsection{Annotation Details} 
\label{app:data_anno}
Raw videos are structured into a three-tier hierarchy: Scene - Task - Subtask by annotators. Each subtask corresponds to the minimal object-centered action unit or a temporally cohesive action sequence.
\noindent \textbf{Atomic (Subtask) Annotation:}
\begin{itemize}
    \item \textbf{Actions (Skills):} Drawn from predefined operation sets (e.g., PICK, PLACE INTO, SEARCH).
    \item \textbf{Target Object:} Semantic structure describing objects (e.g., [cloth, container]).
    \item \textbf{Temporal Bounds:} Precise start/end timestamps delimiting the execution segment in the video.
    \item \textbf{Status:} Labeled as {success, failure, irrelevant} to denote execution outcome.
\end{itemize}

\noindent \textbf{Perception annotation:} We use vision foundation models~\cite{kirillov2023segany,wen2025foundationstereo} to pre-annotate scene elements, followed by manual selection and correction. 
This includes generating instance masks by applying SAM~\cite{kirillov2023segany} to human-provided prompt points and estimating scene depth from first-person binocular images using the FoundationStereo~\cite{wen2025foundationstereo} model.

\noindent \textbf{Vision-language annotation} is generated through manual curation and includes several components. 
\begin{itemize}
    \item \textbf{CoT Reasoning:} We use Qwen2.5-VL-72B~\cite{Qwen2.5-VL} to generate a thought process indicating the position of each subtask within a complete task and when to enter the next subtask for long-horizon task decomposition, where generated thought data is reviewed and corrected by humans. 
    \item \textbf{Spatial Referring (SR):} We select key frames near the action start where the operation target object is clearly visible for some subtasks and add spatial relation terms based on the atomic instruction to uniquely refer to a single target area in the scene.
    \item \textbf{Subtask Prediction (SP):} We select tasks containing 4 to 5 subtasks and use the next subtask after the current one as the correct option, while generating three logically reasonable incorrect options based on the task description.
    \item \textbf{Completion Verification (CV):} We randomly crop 0\% to 35\% of each subtask for a task and use the atomic instruction of each subtask, as well as the cropped fragment, as input to label the task progress status regarding whether the task is completed.   
\end{itemize}
Notably, subtask prediction, spatial referring, and progress verification are annotated on only a subset of the whole WIYH data to construct an evaluation set for validating the capabilities of existing VLMs in human-centric scenarios.


\subsection{Additional Dataset Visualizations}
\begin{figure*}[]
  \centering
    \includegraphics[width=\textwidth]{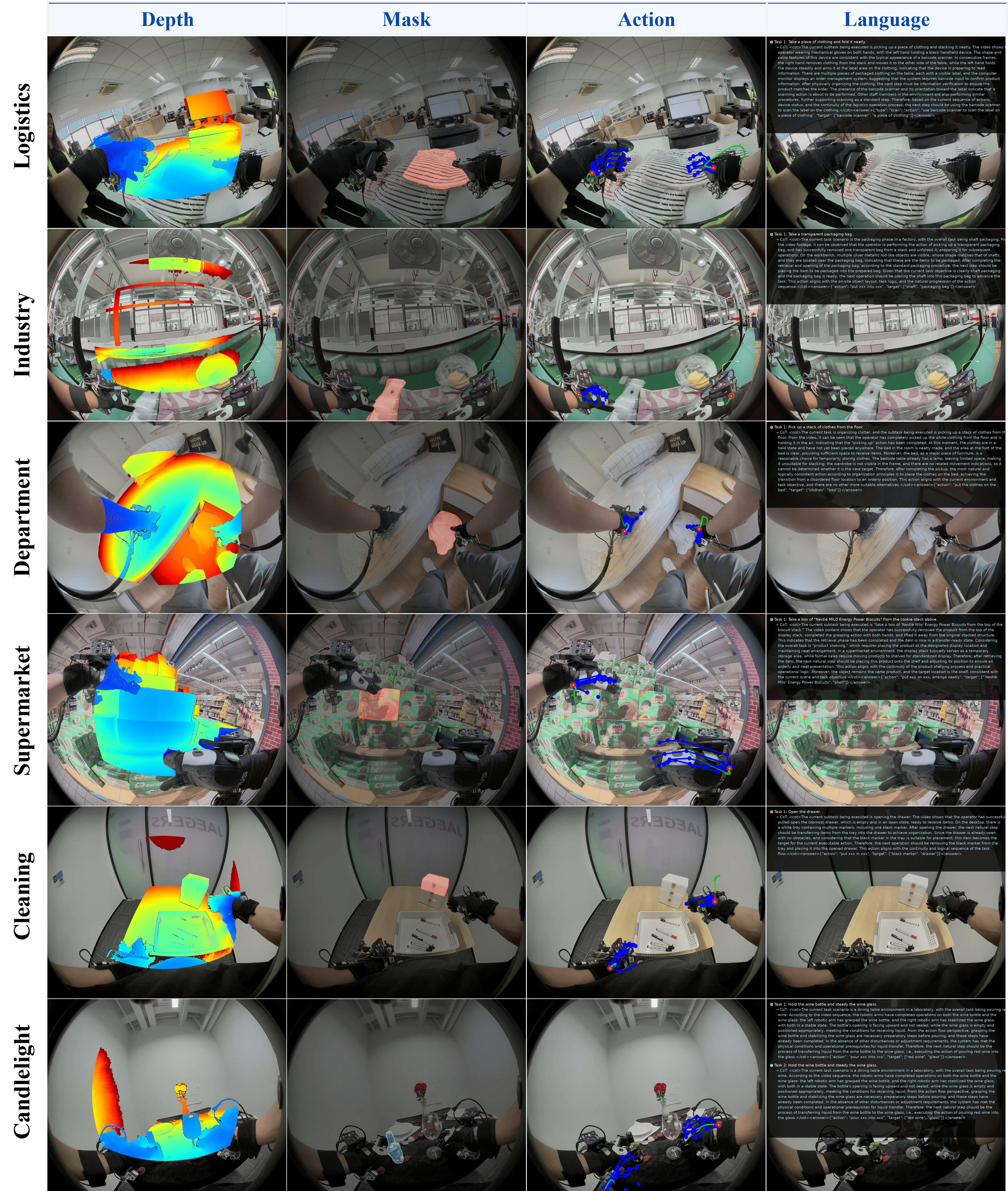}
  \caption{Data annotation samples cross different scenes. The example of ego-centric data annotations, including depth, mask, action, and task descriptions, in six different scenarios.}
  \label{fig:more_vis}
\end{figure*}

\begin{figure*}[]
  \centering
  \includegraphics[width= 0.9\textwidth]{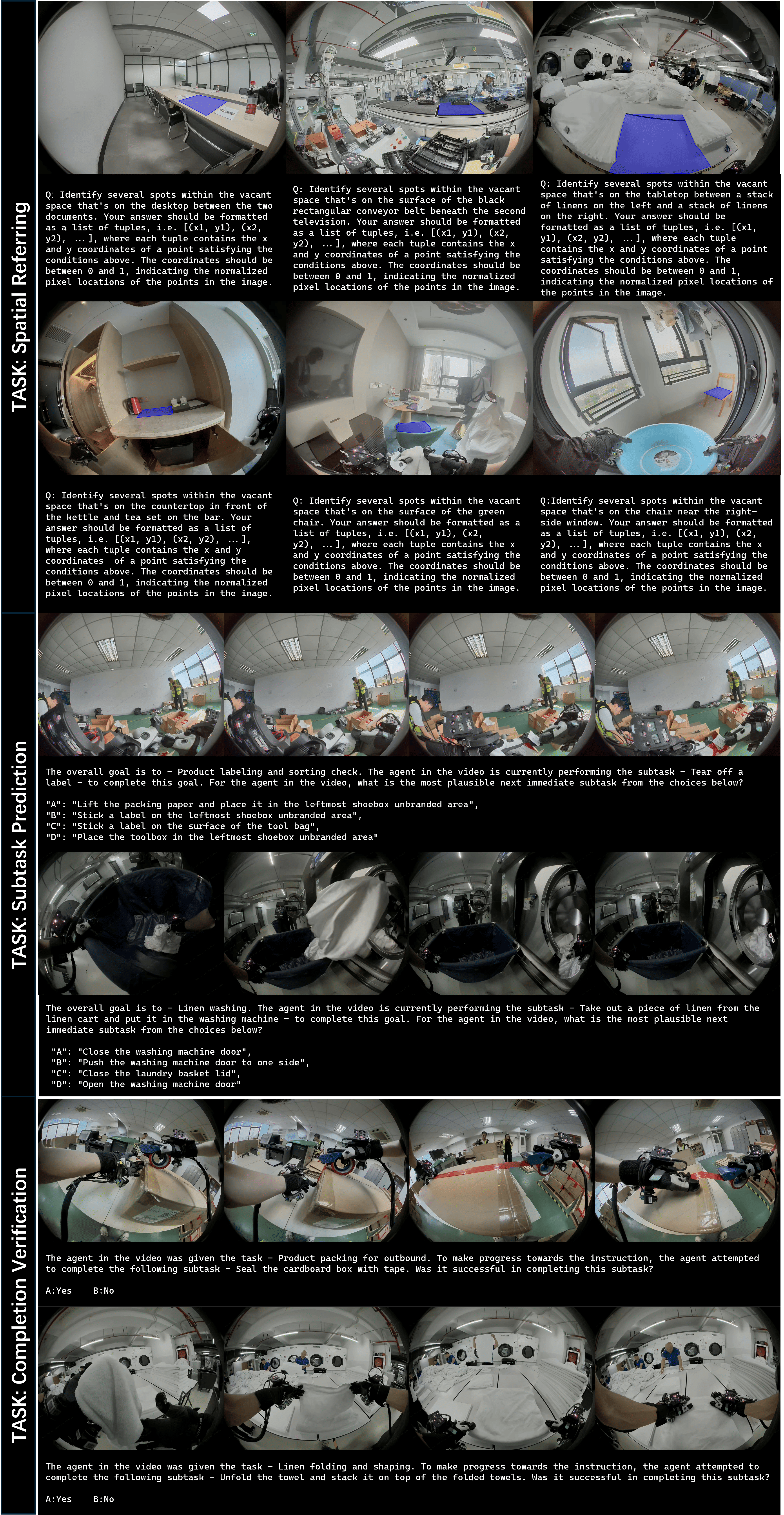}
  \caption{Examples of annotations in the Human-centric Vision-Language Benchmark.}
  \label{fig:vlbenchmark}
\end{figure*}

We present more visualizations about the annotations of our dataset in Figure~\ref{fig:more_vis}. Annotations of the Human-centric Vision-Language tasks are shown in Figure~\ref{fig:vlbenchmark}.

\section{Additional Benchmarks and Applications}
\subsection{Human-centric Vision-Language Benchmark}
\label{benchmark}

The Human-centric Vision-Language (HVL) benchmark is built from the vision-language annotations introduced in Section~\ref{sec:data_collection_annotation}. It evaluates whether existing Multimodal Large Language Models (MLLMs) can understand in-the-wild human manipulation from three complementary perspectives: spatial grounding, long-horizon subtask reasoning, and task-progress verification. All models are evaluated directly without task-specific training, including GPT-4o~\cite{openai2024gpt4o}, Qwen3-VL-Plus~\cite{qwen2025vl}, Doubao-Seed-1.6-Vision~\cite{doubao2024seed}, and Qwen3-VL-4B-Instruct~\cite{qwen2025vl_4b}.

Table~\ref{tab:vlm} summarizes the benchmark results. Existing VLMs achieve relatively high scores on Subtask Prediction, indicating that they can often infer high-level procedural structure from human manipulation videos. In contrast, Spatial Referring remains difficult for all models, and Completion Verification is only slightly above random guessing. These results suggest that general-purpose VLMs still have limited capability in fine-grained embodied spatial understanding and dynamic progress modeling, even when they perform reasonably well on task-level decomposition.

\begin{table}[t]
\centering
\setlength{\tabcolsep}{0.02\linewidth}
\caption{Comparison of VLMs on spatial referring, subtask prediction, and completion verification tasks.}
\resizebox{0.8\linewidth}{!}{
\begin{tabular}{
  l | 
  S[table-format=2.2] 
  S[table-format=2.2] 
  S[table-format=2.2]
}
\toprule

\textbf{Model} & 
{\textbf{S.R. (\%)}} & 
{\textbf{S.P. (\%)}} & 
{\textbf{C.V. (\%)}} \\
\midrule
GPT-4o~\cite{openai2024gpt4o} & 10.04 & 73.40 & 51.18 \\
Qwen3-VL-Plus~\cite{qwen2025vl} & 46.53 & 71.75 & 55.89 \\
Doubao-Seed-1.6-Vision~\cite{doubao2024seed} & 39.83 & 76.84 & 51.96 \\
Qwen3-VL-4B-Instruct~\cite{qwen2025vl_4b} & 25.58 & 69.21 & 56.73 \\
\bottomrule
\end{tabular}
}
\label{tab:vlm}
\end{table}

\subsubsection{Task Definitions and Evaluation Protocol}
\begin{figure}[]
  \centering
  \includegraphics[width=1.0\textwidth]{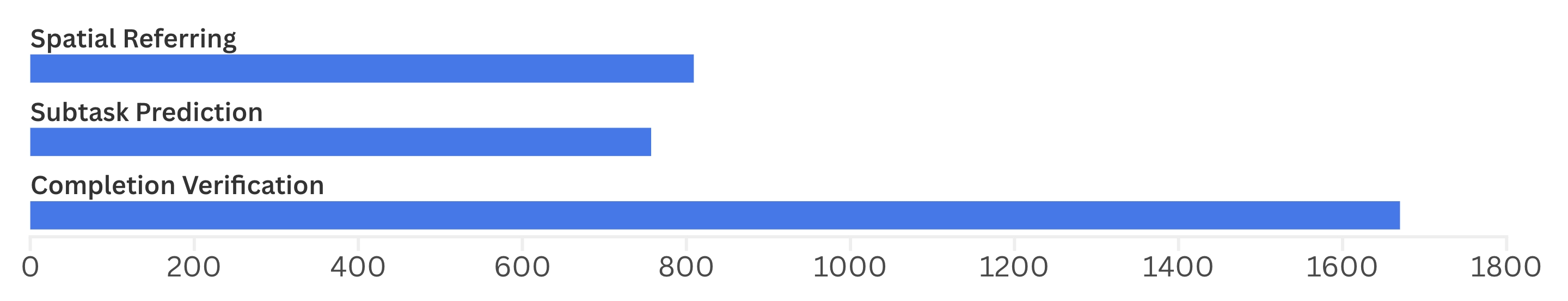}
  \caption{Amount of three task annotations in the Human-centric Vision-Language Benchmark.}
  \label{fig:task_count}
\end{figure}

We sample raw WIYH data and manually annotate three HVL tasks: Spatial Referring, Subtask Prediction, and Completion Verification. Figure~\ref{fig:task_count} reports the amount and distribution of annotations for these tasks.

(1) \textbf{Spatial Referring (S.R.)} evaluates spatial grounding in human manipulation scenes. Given relevant image frames and a question referring to a target region, annotators mark the ground-truth ROI with polygons. The VLM is prompted to output a set of points $P_a$ that correspond to the referred region. We denote by $P_r$ the subset of predicted points that fall inside the ground-truth region and compute:
\begin{equation}
    score_{SR} = \frac{N_r}{N_a},
\end{equation}
where $N_a$ is the number of predicted points in $P_a$, and $N_r$ is the number of points in $P_r$. The score lies in $[0, 1]$, with higher values indicating better spatial referring ability in ego-centric scenarios.

(2) \textbf{Subtask Prediction (S.P.)} evaluates long-horizon task decomposition. Given a human manipulation video clip and the current task description, the model selects the correct next subtask from four candidates. We use Qwen2.5-VL-72B~\cite{Qwen2.5-VL} to generate nine negative candidates based on the video and ground-truth next action, covering both action and target-object variations. Human annotators then select three plausible negatives that fit the scenario. Together with the correct answer, they form a four-option question, for which random selection has an expected accuracy of 25\%.

(3) \textbf{Completion Verification (C.V.)} evaluates task-progress understanding. Given a video segment that is randomly truncated near the end of a subtask and the corresponding task description, the model determines whether the subtask has been completed. We report binary selection accuracy, where random guessing has an expected accuracy of 50\%.

\subsubsection{Tasks Visualization.} 

Figure~\ref{fig:case_study} shows representative samples and annotations for the Spatial Referring task.

\subsubsection{Case Analysis.} 

\begin{figure*}[]
  \centering
  \includegraphics[width=1.0\textwidth]{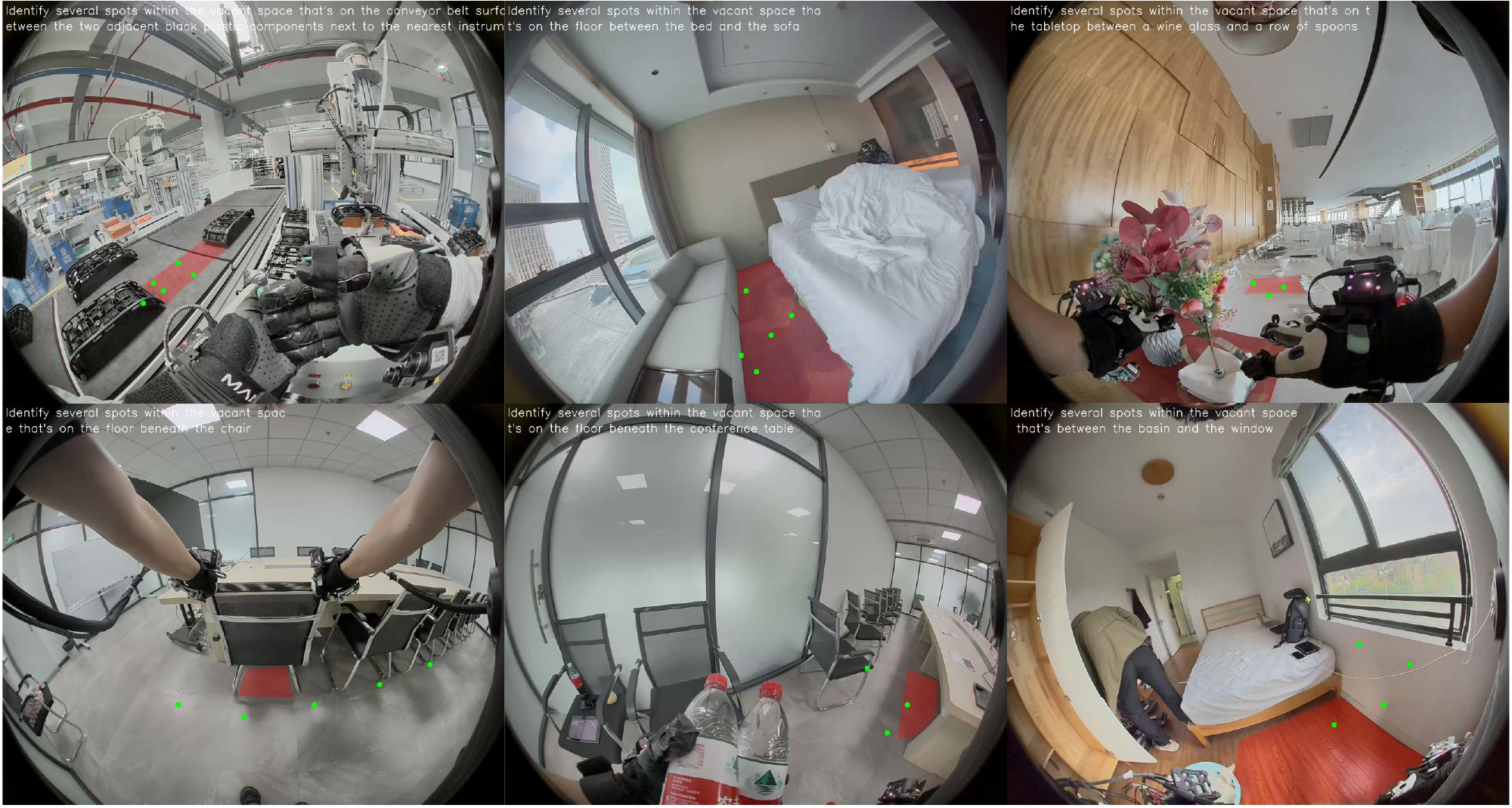}
  \caption{Case analysis of the Spatial Referring task.}
  \label{fig:case_study}
\end{figure*}

Figure~\ref{fig:case_study} presents Qwen3-VL results on Spatial Referring. The model handles some in-plane spatial relations, such as ``between A and B'', but struggles with relations that require reasoning over different height levels, such as ``beneath'' or cases where the two reference objects are located at different depths. This failure mode is consistent with the low S.R. scores in Table~\ref{tab:vlm} and highlights the remaining gap between general visual-language understanding and embodied spatial reasoning.

\subsection{Human-centric World Modeling}

\subsubsection{Dynamic 4D Reconstruction with Gaussian Splatting}
\noindent  \textbf{Evaluation Metrics.} In terms of metrics of 4D reconstruction, we evaluate the RGB rendering quality using PSNR, SSIM, and LPIPS. For PSNR and SSIM, higher is better, while for LPIPS, lower is better. For geometry accuracy from rendered depth maps, we use A.R. (lower is better) and $\delta_1$ (higher is better).

\begin{itemize}
    \item {PSNR (Peak Signal-to-Noise Ratio)},
    $$
    \mathrm{PSNR} = 10 \cdot \log_{10}\!\left( \frac{\mathrm{MAX}_I^2}{\mathrm{MSE}} \right)
    $$
    $$
    \mathrm{MSE} = \frac{1}{MN}\sum_{i=1}^{M}\sum_{j=1}^{N}\big(I(i,j) - K(i,j)\big)^2,
    $$
    where $I$ is the reference image, $K$ is the rendered image, and $\mathrm{MAX}_I$ is the maximum possible pixel value. A higher PSNR indicates that the rendered image is closer to the ground truth.

    \item {SSIM (Structural Similarity Index Measure).}
    $$
    \mathrm{SSIM}(x,y) = 
    \frac{(2\mu_x\mu_y + c_1)\,(2\sigma_{xy} + c_2)}
         {(\mu_x^2 + \mu_y^2 + c_1)\,(\sigma_x^2 + \sigma_y^2 + c_2)},
    $$
    where $\mu$, $\sigma$, and $\sigma_{xy}$ denote the mean, variance, and covariance of image patches $x$ and $y$, respectively. SSIM evaluates perceptual similarity in terms of luminance, contrast, and structure; values closer to $1$ indicate better similarity.

    \item {LPIPS (Learned Perceptual Image Patch Similarity).} 
    This metric computes the perceptual distance between rendered and ground-truth images by extracting deep features from a pretrained CNN and measuring the distance in that feature space. Lower LPIPS values indicate that two images are more perceptually similar.

    \item {AbsRel (Absolute Relative Error).}
    $$
    \mathrm{AbsRel} = \frac{1}{N} \sum_{i=1}^{N} \frac{\lvert d_i - d_i^* \rvert}{d_i^*},
    $$
    where $d_i$ is the predicted depth at the pixel $i$ and $d_i^*$ is the corresponding ground-truth depth. A lower value denotes a more accurate geometry.

    \item {$\delta_1$ (Threshold Accuracy / Delta1).}
    $$
    \delta_1 = \frac{1}{N} \sum_{i=1}^{N} 
    \mathbf{1}\!\left( \max\!\left(\frac{d_i}{d_i^*},\,\frac{d_i^*}{d_i}\right) < 1.25 \right),
    $$
    which measures the percentage of predicted depth values that are within a factor of $1.25$ of the ground-truth depth. A higher value $\delta_1$ indicates better accuracy.
\end{itemize}

During optimization, all input videos are processed at a resolution of 960×720. We represent each scene with 40k dynamic Gaussians and 100k static Gaussians and use B = 10 shared SE(3) motion bases for all points. We optimize the model with Adam, running 1k iterations for the initialization stage and 500 training epochs, with learning rates of $1.6 \times 10^{-4}$ for Gaussian means and SE(3) bases, $5 \times 10^{-3}$ for scales, $1 \times 10^{-3}$ for rotations, and $1 \times 10^{-2}$ for colors, opacities, and motion coefficients. Training on a 300-frame sequence takes about 2 hours on a single A100 GPU and supports real-time rendering at around 140 FPS.

\subsubsection{Language-conditioned Video Generation}

This section evaluates the contribution of our proposed WIYH dataset to enhancing video generation performance in ego-centric manipulation tasks. 
We fine-tuned two baseline models, transformer-based CogVideo~\cite{hong2022cogvideo} and UNet-based DynamiCrafter~\cite{xing2024dynamicrafter}, initialized with their original weights, using WIYH under the image-and-text-to-video setting. During fine-tuning, all input videos were resized to 480p resolution. We randomly select 90\% of the data as the training set and the remaining 10\% as the test set.

\begin{figure}[]
  \centering
  \setlength{\abovecaptionskip}{0.cm}
  \includegraphics[width=0.9\textwidth]{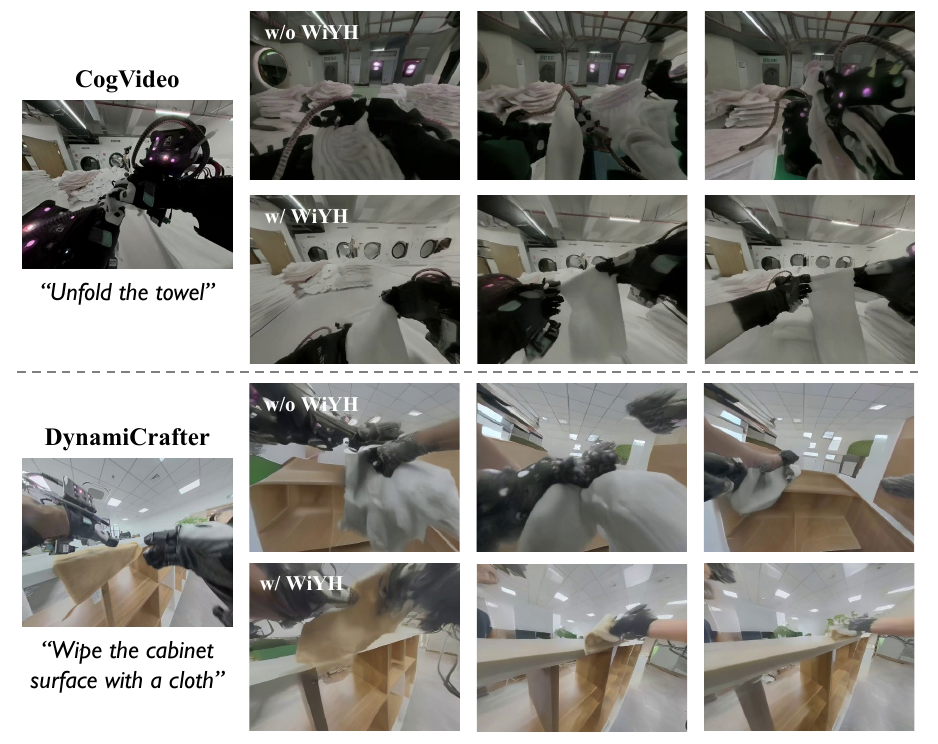}
  \caption{\textbf{Language-Conditioned Video Generation.} When provided with language instructions, two baseline video prediction methods exhibited significant hallucinations without WIYH fine-tuning. However, after fine-tuning on our dataset, they demonstrated a markedly enhanced ability to imagine future states.}
  \label{fig:video_gen}
\end{figure}

As shown in Figure~\ref{fig:video_gen}, models fine-tuned with WIYH generate egocentric videos with improved spatial and temporal coherence, as well as stronger alignment with both the input image and action instructions. For example, in the “Wipe the cabinet” task, the cabinet’s orientation and the towel’s color are reproduced more accurately. In contrast, videos generated by non-fine-tuned models exhibit noticeable object fragmentation and distortion, along with discontinuous and physically implausible hand–object interactions. Notably, after fine-tuning, the complex motions of non-rigid objects in the generated videos appear more realistic and natural. These improvements demonstrate that WIYH substantially enhances a model’s ability to capture real-world dynamics.

Quantitative results are summarized in Table~\ref{tab:video_gen}. We use four metrics from VBench~\cite{huang2024vbench}—consistency, smoothness, dynamic (motion intensity), and quality in order to assess motion plausibility, temporal coherence, and visual fidelity of the video generation models. Following VBench~\cite{huang2024vbench}, the evaluation metrics are as follows:
\begin{itemize}
    \item \textbf{Consistency.} To evaluate the temporal consistency of the target manipulation subjects, we calculate the DINO~\cite{caron2021emerging} feature similarity across frames. For assessing the temporal consistency of the background scenes, we compute the CLIP~\cite{radford2021learning} feature similarity across frames. The overall consistency score is defined as the arithmetic mean of the subject consistency and the background consistency.
    \item \textbf{Smoothness.} To assess the temporal continuity of the generated videos, we employ a motion-consistency metric based on the AMT~\cite{li2023amt} video frame interpolation model. Given a generated video with frames \( \{f_0, f_1, \ldots, f_{2n-1}, f_{2n}\} \), we first remove all odd-indexed frames to obtain \( \{f_0, f_2, \ldots, f_{2n}\} \). The AMT model is then used to interpolate the missing frames, producing \( \{\hat{f}_1, \hat{f}_3, \ldots, \hat{f}_{2n-1}\} \). Finally, we compute the mean absolute error between the interpolated and original frames, where a larger value indicates smoother and more temporally coherent motion.
    \item \textbf{Dynamic.} To prevent the model from producing static videos in pursuit of higher temporal consistency scores, we employ RAFT~\cite{teed2020raft} to estimate optical flow strengths between consecutive frames of each generated video. The average of the largest 5\% optical flows is used to determine whether the video is static. The final dynamic score is calculated by measuring the proportion of non-static videos generated by the model.
    \item \textbf{Quality.} To assess the visual quality of generated video frames, we employ the MUSIQ~\cite{ke2021musiq} image quality predictor trained on the SPAQ~\cite{fang2020perceptual} dataset. This metric quantifies common distortions such as overexposure, noise, and blur, with a higher score indicating fewer artifacts.
\end{itemize}
After fine-tuning with the WIYH dataset, both baseline models show clear improvements across all metrics, with the most significant gains observed in dynamics and temporal consistency.

\begin{table}[b]
\setlength{\tabcolsep}{0.027\linewidth}
\centering
\caption{Comparison of consistency, smoothness, dynamics, and quality metrics under different conditions.}
\footnotesize
\resizebox{0.9\columnwidth}{!}{ 
\begin{tabular}{
  c | 
  S[table-format=2.1] 
  c 
  S[table-format=2.1] 
  c 
}
\toprule
& \multicolumn{2}{c}{\textbf{CogVideo}} 
& \multicolumn{2}{c}{\textbf{DynamiCrafter}} \\ 
\cmidrule(lr){2-3} \cmidrule(lr){4-5} 
\textbf{Metric} & {\textbf{w/o WIYH}} & {\textbf{w/ WIYH}} & {\textbf{w/o WIYH}} & {\textbf{w/ WIYH}} \\
\midrule
Consistency $\uparrow$ & 80.6 & 88.2 \textbf{(\textcolor{green}{+7.6})} & 85.3 & 91.5 \textbf{(\textcolor{green}{+6.2})} \\
Smoothness $\uparrow$  & 93.5 & 98.4 \textbf{(\textcolor{green}{+4.9})} & 97.4 & 98.6 \textbf{(\textcolor{green}{+1.2})} \\
Dynamic $\uparrow$     & 62.9 & 78.5 \textbf{(\textcolor{green}{+15.6})} & 74.8 & 89.6 \textbf{(\textcolor{green}{+14.8})} \\
Quality $\uparrow$     & 60.1 & 67.6 \textbf{(\textcolor{green}{+7.5})} & 63.5 & 70.6 \textbf{(\textcolor{green}{+7.1})} \\
\bottomrule
\end{tabular}
} 
\label{tab:video_gen} 
\end{table}

\section{Robot Policy Evaluation Details}
\label{app:policy_eval_details}

\subsection{Task-matched Ego-Robot Co-training Details}
\label{app:cotrain_details}

In the task-matched ego-robot co-training study, we construct a combined training set from two sources: robot demonstrations collected in constrained scenes and ego-centric demonstrations of the same task collected with Oracle Suite. The robot system uses a 6-DoF dexterous hand (ROH-LiteS001) as the end-effector, and the hand trajectories from ego-centric demonstrations are retargeted~\cite{li2025h2r} to match the hand degrees of freedom. Robot demonstrations start from a limited set of initial poses in single-object scenes, while ego-centric demonstrations are collected in cluttered multi-object scenes and provide broader observation and action coverage.

\begin{figure*}[]
  \centering
  \setlength{\abovecaptionskip}{0.cm}
  \includegraphics[width= \textwidth]{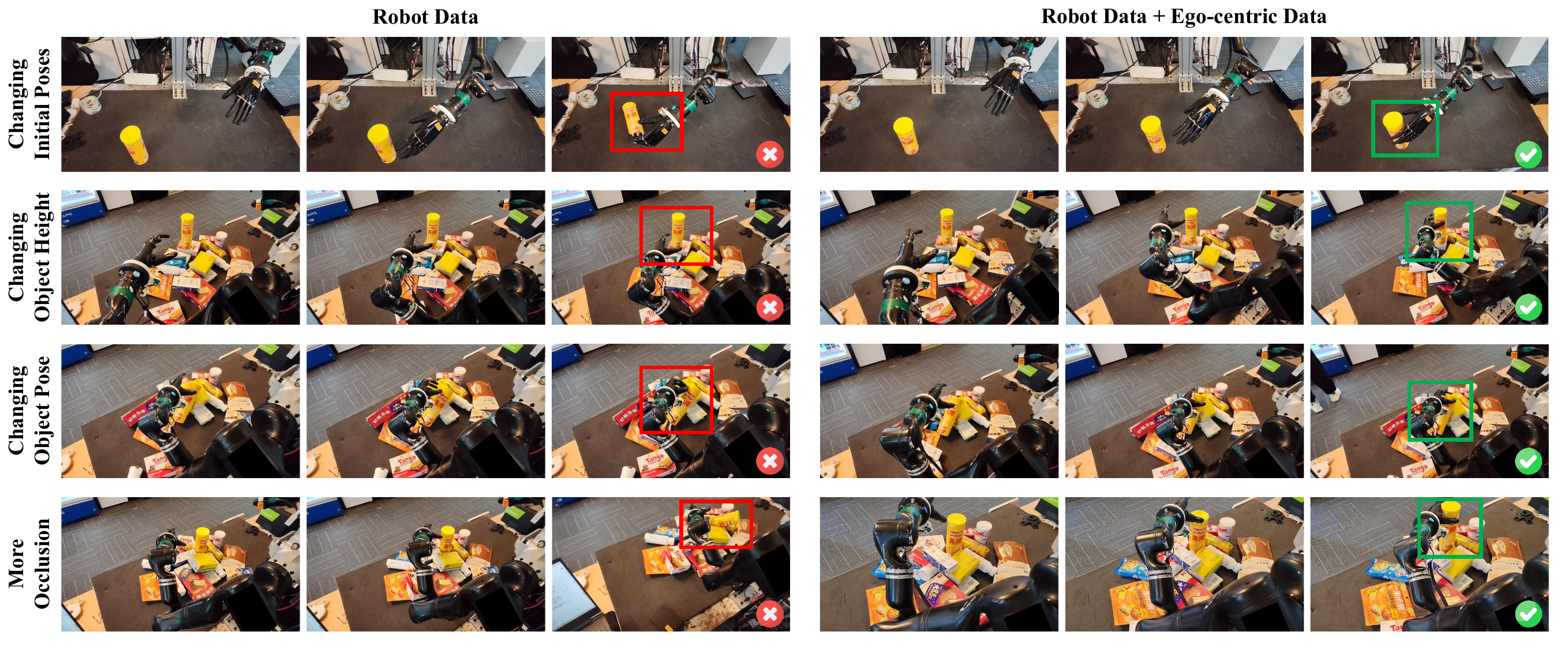}
  \caption{\textbf{Real-robot manipulation experiments.} We present the comparison of manipulation policy performance under four novel task settings, where the policies are trained either exclusively on robot data collected or co-trained using both robot data and annotated ego-centric data.}
  \vspace{-1mm}
  \label{fig:real_exp}
\end{figure*}

\begin{figure*}[]
  \centering
  \includegraphics[width= 0.88 \textwidth]{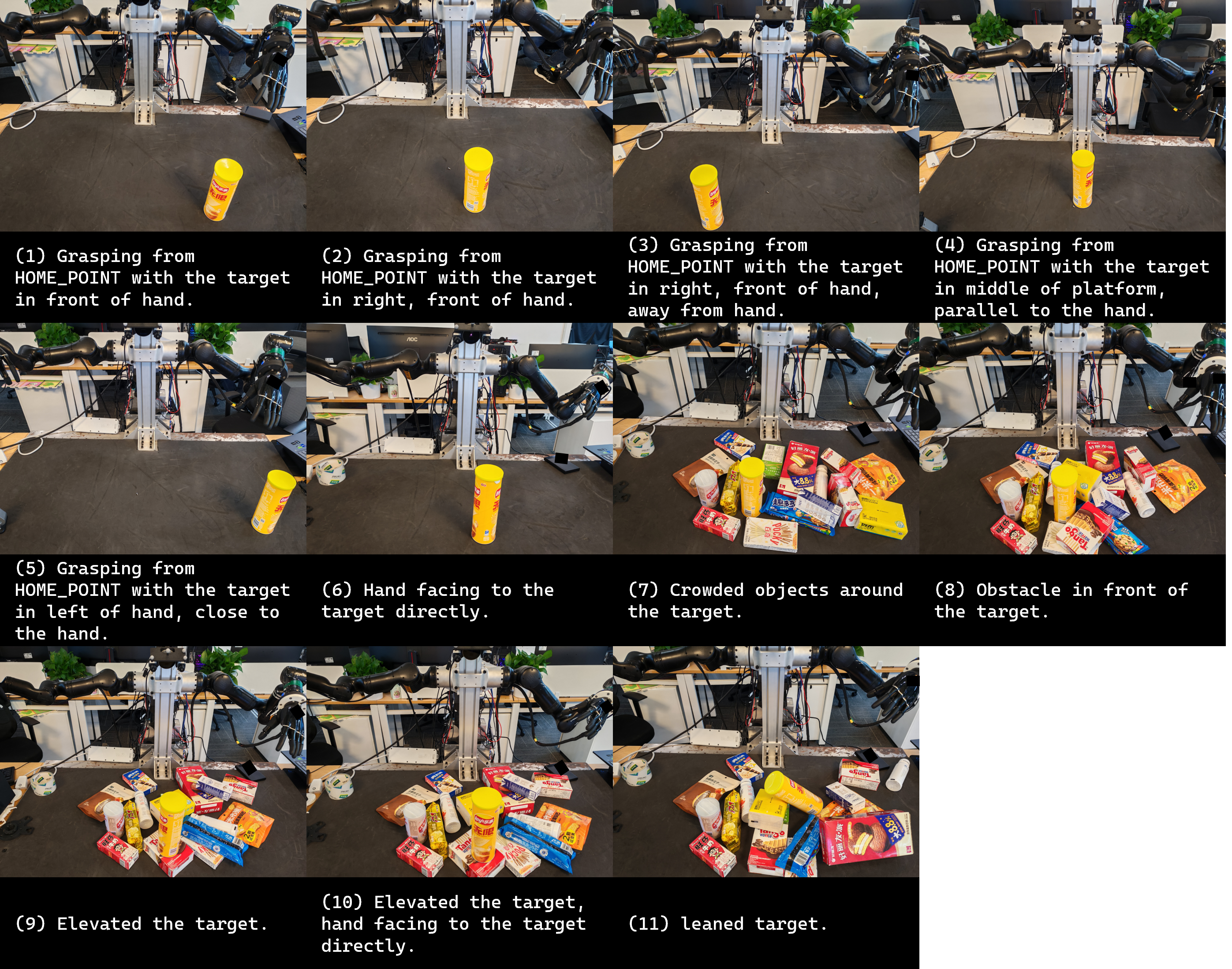}
  \caption{Task variant visualization in the co-training experiment.}
  \label{fig:ce_tasks}
\end{figure*}

\begin{figure*}[]
  \centering
  \includegraphics[width= 1.0\textwidth]{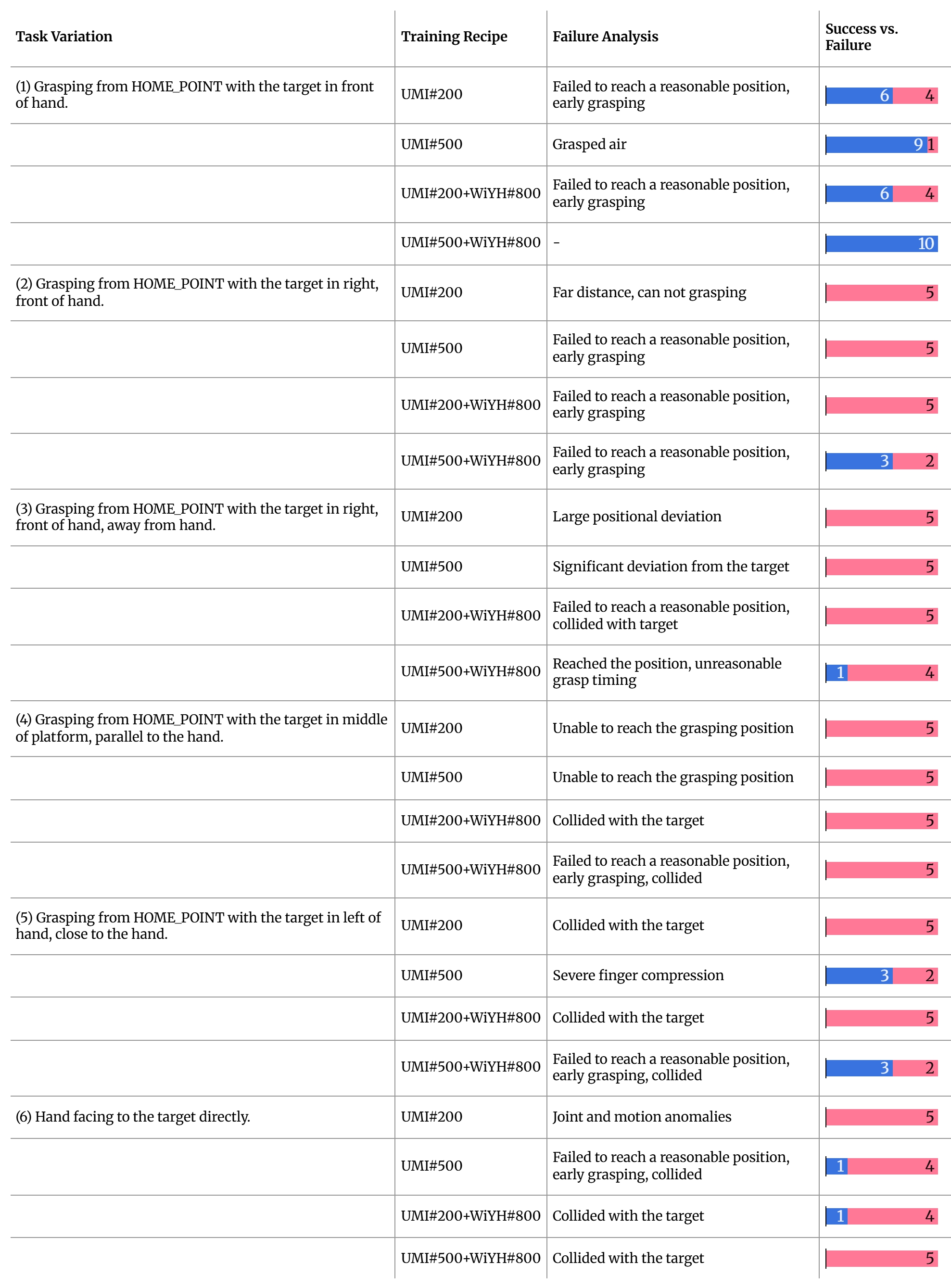}
  \caption{Dexterous manipulation Details - 1.}
  \label{fig:exp_table1}
\end{figure*}

\begin{figure*}[]
  \centering
  \includegraphics[width= 1.0\textwidth]{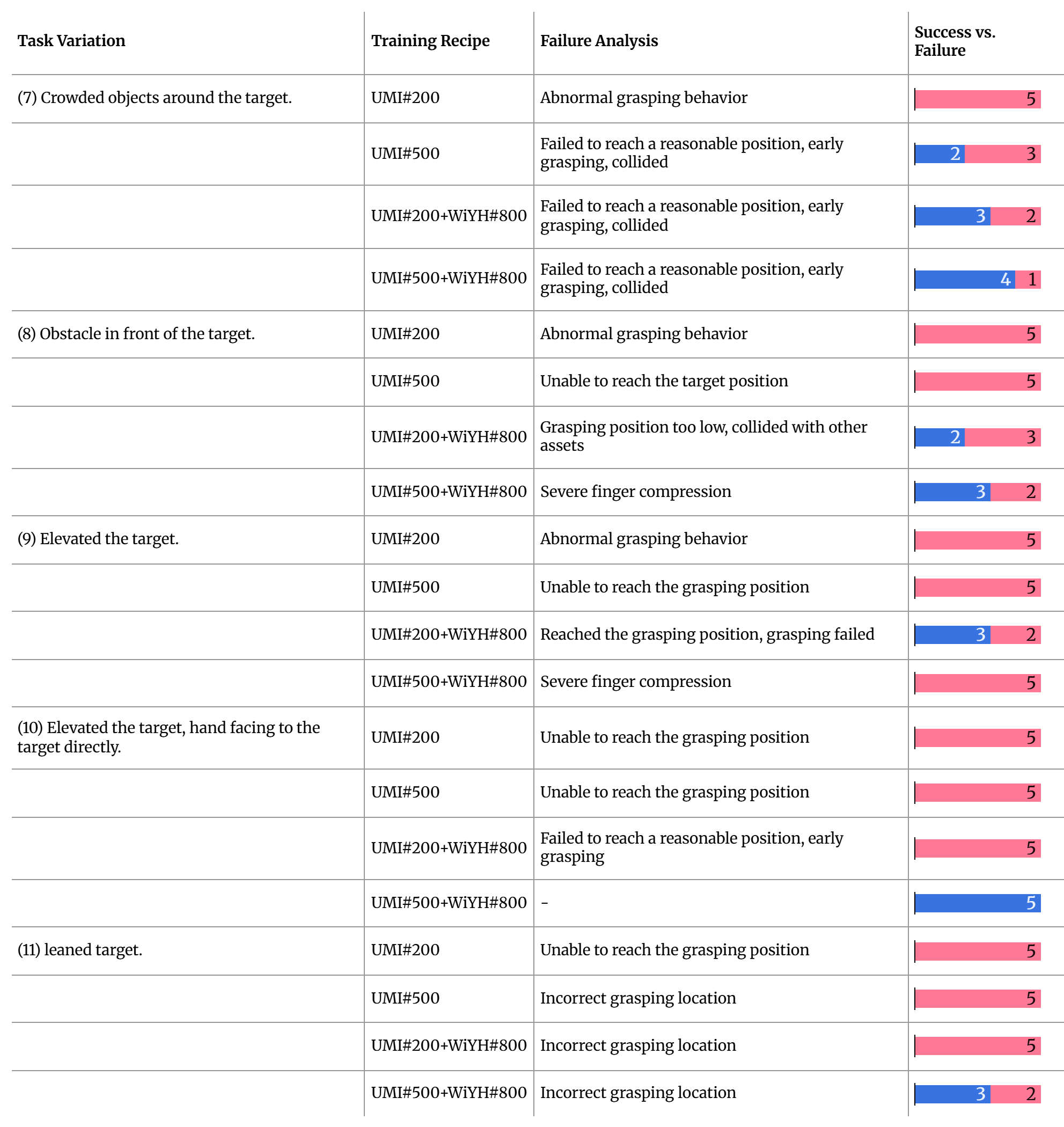}
  \caption{Dexterous manipulation Details - 2.}
  \label{fig:exp_table2}
\end{figure*}

As shown in Figure~\ref{fig:ce_tasks}, we evaluate 11 task variants under two scenarios: Single-Object Scene and Multi-Object Cluttered Scene. The Single-Object Scene evaluates generalization to randomized robot initial poses that differ from those used during robot-data collection. The Multi-Object Cluttered Scene evaluates generalization under varying scene height, target-object poses, object layouts, and occlusions. Figure~\ref{fig:real_exp} shows qualitative results. Figures~\ref{fig:exp_table1} and~\ref{fig:exp_table2} report the success rates and failure causes for each task in the two scenarios.

As shown in Figures~\ref{fig:exp_table1} and~\ref{fig:exp_table2}, incorporating ego-centric human data generally improves success rates. In the Single-Object Scene, the diverse action distribution of human data improves generalization to unseen initial states, enabling successful grasping from different positions and longer distances. In the Multi-Object Cluttered Scene, policies trained only on robot data show limited generalization under changes in object height, layout, and occlusion, whereas adding human data introduces broader action and observation coverage. The consistently low success rates on some tasks also point to future directions, including improving the precision of human-data collection and designing more effective algorithms for transferring human demonstrations to dexterous robot manipulation.

\subsection{RoboCasa Pre-training Details}
\label{app:robocasa_details}

For the RoboCasa pre-training study, we compare ego-centric pre-training using EgoDex, VITRA, WIYH, and WIYH*. WIYH* denotes an unverified variant of WIYH that does not use the reprojection-based filtering or manual task-segmentation verification described in Section~\ref{sec:check}, and therefore contains noisier action labels and less reliable action-language alignment. To isolate data quality from dataset scale, all pre-training datasets are matched to 80K episodes. We use Fast-WAM~\cite{yuan2026fast} as the base policy model and pre-train each model for 30K steps. For post-training, we use 10\% of the official RoboCasa demonstrations for each task, creating a low-robot-data regime in which performance remains sensitive to pre-training quality. Each task is evaluated with 50 trials.

RoboCasa consists of 24 distinct tasks, categorized into articulated-object tasks and pick-and-place tasks. The articulated-object tasks include: \textit{PnP Bottle To Cabinet Close}, \textit{PnP Can To Drawer Close}, \textit{PnP Cup To Drawer Close}, \textit{PnP Milk To Microwave Close}, \textit{PnP Potato To Microwave Close}, and \textit{PnP Wine To Cabinet Close}. The pick-and-place tasks include: \textit{PnP Novel From Cuttingboard To Basket}, \textit{PnP Novel From Cuttingboard To Cardboardbox}, \textit{PnP Novel From Cuttingboard To Pan}, \textit{PnP Novel From Cuttingboard To Pot}, \textit{PnP Novel From Cuttingboard To Tieredbasket}; \textit{PnP Novel From Placemat To Basket}, \textit{PnP Novel From Placemat To Bowl}, \textit{PnP Novel From Placemat To Plate}, \textit{PnP Novel From Placemat To Tieredshelf}; \textit{PnP Novel From Plate To Bowl}, \textit{PnP Novel From Plate To Cardboardbox}, \textit{PnP Novel From Plate To Pan}, \textit{PnP Novel From Plate To Plate}; and \textit{PnP Novel From Tray To Cardboardbox}, \textit{PnP Novel From Tray To Plate}, \textit{PnP Novel From Tray To Pot}, \textit{PnP Novel From Tray To Tieredbasket}, and \textit{PnP Novel From Tray To Tieredshelf}.

\subsection{Real-Robot Transfer Details}
\label{app:realrobot_details}

For real-robot transfer, we pre-train the policy on WIYH data with atomic instruction annotations and post-train it on two real-world tasks: rose insertion and gift packing. For each task, we collect 150 gripper-based teleoperation demonstrations for post-training. Each setting is evaluated for 10 trials per task under varied object positions and target objects, including seen and unseen bottles for rose insertion and seen and unseen gifts for gift packing.

\section{Ethics and Privacy Statement}
Our work involves large-scale data collection of real-world human manipulation activities. We took a number of measures to ensure responsible data handling, privacy protection, and ethical compliance.

\noindent \textbf{Data Collection and Consent.}
All human subjects participating in data collection were recruited voluntarily and provided informed consent in accordance with institutional guidelines. Participants were briefed on the purpose of the study, the modalities collected (RGB video, depth, motion capture), the intended research use, and their right to withdraw at any time.

\noindent \textbf{Privacy Protection.}
Although the dataset contains real-world human activities and may capture visible human faces or identifiable biometric patterns, several privacy-preserving strategies were implemented:
\begin{itemize}
    \item \textbf{Controlled subjects only}: No bystanders or non-consenting individuals appear in the dataset; all scenes were recorded in controlled environments with only consenting operators present.
    \item \textbf{No identity labels}: We do not include personal information, demographic information, or identity annotations.
    \item \textbf{Face and identity protection}: When faces are visible in RGB streams, we apply optional face blurring in the public release version.
    \item \textbf{No audio recordings}: We do not collect speech or audio signals to avoid unintended disclosure of personal information.
\end{itemize}

\noindent \textbf{Sensitive Content and Safety.}
The dataset does not include minors, hazardous behaviors, or sensitive activities. All recordings were conducted in standard workplace or daily-life environments using safety protocols.

\noindent \textbf{Data Usage and Licensing.}
The released dataset is intended exclusively for research on embodied AI, manipulation, perception, and VLA learning. Redistribution or commercial use is restricted according to the license accompanying the dataset. Users are required to comply with ethical research practices and avoid any attempts at identity recognition or misuse of human-related data.

\noindent \textbf{Potential Risks and Mitigations.}
As with any dataset containing human recordings, there is a theoretical risk of model misuse, such as identity extraction or surveillance applications. To mitigate these risks, we provide:
(1) no textual identity metadata,
(2) optional visual anonymization tools, and
(3) a license that explicitly forbids re-identification, face recognition, or surveillance-related uses.

\noindent \textbf{Institutional Review.}
All data collection procedures were reviewed and approved by an internal ethics committee to ensure compliance with privacy protection and responsible AI principles.

\end{document}